\documentclass[letterpaper]{article} 
\usepackage{aaai2026}  
\usepackage{times}  
\usepackage{helvet}  
\usepackage{courier}  
\usepackage[hyphens]{url}  
\usepackage{graphicx} 
\usepackage{natbib}  
\usepackage{caption} 
\usepackage{algorithm}
\usepackage{algorithmic}

\usepackage{newfloat}
\usepackage{listings}
\DeclareCaptionStyle{ruled}{labelfont=normalfont,labelsep=colon,strut=off} 
\floatstyle{ruled}
\newfloat{listing}{tb}{lst}{}
\floatname{listing}{Listing}
\usepackage{graphicx}	
\usepackage{amsmath}	
\usepackage{amssymb}	
\usepackage{booktabs}
\usepackage{times}
\usepackage{microtype}
\usepackage{epsfig}
\usepackage{caption}
\usepackage{float}
\usepackage{placeins}
\usepackage{color, colortbl}
\usepackage{tabularx}
\usepackage{xstring}
\usepackage{multirow}
\usepackage{xspace}
\usepackage{url}
\usepackage{subcaption}
\usepackage{xcolor}
\usepackage{adjustbox}
\usepackage{makecell}
\usepackage{amsthm}

\newtheorem{lemma}{Lemma}
\newtheorem{proposition}{Proposition}

\newtheorem{definition}{Definition}

\title{CAD-VAE: Leveraging Correlation-Aware Latents \\for Comprehensive Fair Disentanglement}
\author{
    Chenrui Ma\textsuperscript{\rm 1},
    Xi Xiao\textsuperscript{\rm 2},
    Tianyang Wang\textsuperscript{\rm 2},
    Xiao Wang\textsuperscript{\rm 3},
    Yanning Shen\textsuperscript{\rm 1}\thanks{Corresponding author: yannings@uci.edu.}
}
\affiliations{
    \textsuperscript{\rm 1}University of California, Irvine, Irvine, CA 92697, USA\\
    \textsuperscript{\rm 2}University of Alabama at Birmingham, Birmingham, AL 35294, USA\\
    \textsuperscript{\rm 3}Oak Ridge National Laboratory, Oak Ridge, TN 37830, USA\\
}

\usepackage{bibentry}

\begin{document}

\maketitle



\begin{abstract}

While deep generative models have significantly advanced representation learning, they may inherit or amplify biases and fairness issues by encoding sensitive attributes alongside predictive features. Enforcing strict independence in disentanglement is often unrealistic when target and sensitive factors are naturally correlated. To address this challenge, we propose \textbf{CAD-VAE} (\textbf{C}orrelation-\textbf{A}ware \textbf{D}isentangled \textbf{VAE}), which introduces a correlated latent code to capture the information shared between the target and sensitive attributes. Given this correlated latent, our method effectively separates overlapping factors without extra domain knowledge by directly minimizing the conditional mutual information between target and sensitive codes. A relevance-driven optimization strategy refines the correlated code by efficiently capturing essential correlated features and eliminating redundancy. Extensive experiments on benchmark datasets demonstrate that CAD-VAE produces fairer representations, realistic counterfactuals, and improved fairness-aware image editing. Source code is available : https://github.com/merry7cherry/CAD-VAE

\end{abstract}    
\section{Introduction}
\label{sec:intro}

Deep generative models have achieved remarkable success in capturing complex data distributions for applications ranging from image synthesis~\cite{bai2024meissonic, ma2025stochastic, chenrui2025Learn} to video generation~\cite{DBLP:conf/nips/MontanaroAAVM24}. In particular, variational autoencoders (VAEs)\cite{higgins2017beta,kingma2022autoencodingvariationalbayes,mathieu2019disentanglingdisentanglementvariationalautoencoders, ma2025beyond} have provided a principled approach to representation learning, where data are encoded into compact latent variables that effectively capture meaningful factors of variation. However, while these latent representations have enabled impressive performance in numerous tasks, concerns about fairness have emerged, as models can inadvertently learn and amplify biases present in training data\cite{Jang_2024_CVPR}.

Such fairness issues arise when the target label and sensitive label become entangled due to societal or dataset biases~\cite{lahoti2020fairnessdemographicsadversariallyreweighted,liu2021justtraintwiceimproving}. To address these problems, existing methods commonly fall into two categories. Invariant learning techniques aim to remove sensitive attributes from the learned representation, often via adversarial training or additional regularization~\cite{lahoti2020fairnessdemographicsadversariallyreweighted,roy2019mitigatinginformationleakageimage}. By contrast, disentanglement approaches encourage the model to partition its latent space into separate codes for target and sensitive information, seeking statistical independence among them~\cite{creager2019flexibly,liu2023fair}. 
Although these solutions have made progress, they typically assume minimal correlation between target and sensitive factors or enforce strict separation via mutual information penalties~\cite{chen2018isolating}. However, multiple works~\cite{Jang_2024_CVPR,park2020readmerepresentationlearningfairnessaware} have demonstrated that achieving fully fair disentanglement is fundamentally impossible under realistic conditions. First, many datasets contain unwanted correlations between the target label and sensitive attributes due to societal bias, making it infeasible to preserve all predictive cues while completely discarding sensitive information~\cite{theaccuracy}. Second, certain features inherently influence both target and sensitive attributes, so perfectly partitioning features into disjoint latent spaces is unachievable without compromising prediction accuracy~\cite{Scalingup}. In these circumstances, any attempt at full disentanglement faces an inevitable trade-off between fairness and utility. 

A natural way to handle this correlation is to explicitly model how target and sensitive attributes overlap. For instance, some methods rely on causal graphs to separate task-relevant features and capture their relationships with sensitive variables~\cite{kim2021counterfactual, sanchez2022vaca, zhu2023learning, hwa2024enforcing}. However, constructing such graphs requires extensive domain knowledge, which is often challenging to acquire in real-world scenarios. 


Motivated by the limitations of existing methods, an correlated latent code is 
introduced to capture the shared information between target and sensitive attributes. 
Our approach advances existing methods~\cite{Jang_2024_CVPR,park2020readmerepresentationlearningfairnessaware} by a directly minimizing conditional mutual information mechanism to achieve disentanglement and an explicit relevance learning strategy to learn the correlated latent code efficiently and properly, as summarized follow:
\begin{enumerate}
    \item We propose a novel correlation-aware representation learning framework that directly minimizes the conditional mutual information between target and sensitive property, conditioned on the correlated latent code, effectively addressing the conflict between predictive objectives and disentanglement.
    \item We introduce an explicit relevance-driven optimization strategy that precisely regulates the correlated latent code, ensuring it captures only the essential shared information without extra domain knowledge.
    \item We validate our approach through comprehensive experiments on multiple benchmark datasets, demonstrating its superiority in achieving correlation-aware disentanglement, enhancing fair prediction performance, and improving both counterfactual generation and fairness-aware image editing, as well as its broad applicability in the context of Vision-Language Models (VLM).
\end{enumerate}

\section{Related Work}
\label{sec:related}

\subsection{Fair Disentanglement Learning} \label{rw:disentangle}
Fair disentanglement methods aim to separate representations into target-related and sensitive-related latent codes rather than directly removing sensitive information \cite{liu2023fair, madras2018learning, xu2018fairgan,10579040}. Early works such as $\beta$-VAE \cite{higgins2017beta} and FactorVAE \cite{kim2018disentangling} introduced mechanisms for semantic decomposition of latent factors, with FactorVAE promoting independence across dimensions by reducing total correlation. Building on this foundation, FairFactorVAE \cite{liu2023fair} further restricts sensitive leakage within the disentanglement process.

Subsequent studies refine these ideas by emphasizing flexible or guided decomposition. FFVAE \cite{creager2019flexibly} adapts latent structures to better isolate sensitive attributes, while GVAE \cite{ding2020guided} employs adversarial constraints to suppress unwanted information. Other strategies incorporate structural priors such as orthogonality in ODVAE \cite{sarhan2020fairness} or distance-covariance minimization in FairDisCo \cite{liu2022fair}. These approaches collectively illustrate progress in disentanglement, yet also reveal the challenge of fully separating target and sensitive information when these factors are inherently correlated.

\subsection{Correlation-Aware Learning} \label{rw:correlation}
Despite advancements in disentanglement, perfect independence between latent codes is difficult to achieve due to natural correlations between sensitive and target attributes \cite{mehrabi2021survey, Jang_2024_CVPR}. For example, facial attributes in CelebA \cite{liu2015deep}, such as ``mustache,'' correlate with both gender and attractiveness, complicating clean separation.

Correlation-aware learning frameworks seek to address this by leveraging causal graphs to categorize latent variables according to their relationships with sensitive attributes \cite{kim2021counterfactual, sanchez2022vaca, zhu2023learning, hwa2024enforcing}. However, causal-graph construction requires strong domain knowledge, and inaccurate assumptions may hinder independence \cite{Jang_2024_CVPR}. FADES \cite{Jang_2024_CVPR} mitigates this dependency by grouping samples across attributes to approximate conditional mutual information and capture shared sensitive-relevant structure. While effective in some contexts, this indirect method may still allow leakage and offers limited control over relevance allocation. These limitations motivate approaches that directly optimize conditional independence with explicit guidance for balancing sensitive-relevant information.

\subsection{Counterfactual Fairness}
Counterfactual fairness (CF) evaluates whether predictions remain stable when sensitive attributes are hypothetically altered \cite{kusner2017counterfactual}. Causal inference is widely used to generate counterfactual instances \cite{zhou2024counterfactual, jung2025counterfactually, zhu2023learning, chiappa2019path, wu2019counterfactual}, enabling comparisons between factual and hypothetical outcomes.

Graph-based CF models \cite{kim2021counterfactual, li2025local} rely on predefined causal structures to produce realistic counterfactuals, but these structures demand precise domain knowledge; inaccurate models may lead to implausible counterfactuals, such as depicting a female subject with a mustache. Although CF offers strong theoretical grounding, its reliance on domain expertise limits practical deployment. To overcome this challenge, our approach introduces a correlated latent code with an explicit relevance-learning mechanism, allowing the model to autonomously learn attribute relationships and enhance counterfactual fairness without external causal assumptions.

\section{Preliminary}
\label{sec:preliminary}

\subsection{Conditional Independence and Mutual Information} \label{pre:ci_mi}


\begin{proposition}[Conditional Independence]
\label{prop:ci_factorization}
Let $A$, $B$, and $C$ be random variables. We say that $A$ is \textbf{conditionally independent} of $B$ given $C$, denoted $A \perp B \mid C$, if and only if their conditional joint probability distribution factorizes as follows:
\vspace{-0.5em}
\begin{equation}
\label{eq:ci_factorization_formula}
p(A, B \mid C) = p(A \mid C) p(B \mid C).
\end{equation}
\vspace{-0.5em}
\end{proposition}
\vspace{-1.5em}

Directly measuring the degree of conditional independence by computing the divergence between the two sides of Eq.~\eqref{eq:ci_factorization_formula} is often intractable in practice, especially in the context of deep learning models where the underlying distributions are complex and high-dimensional. Instead, a common and more tractable approach is to use an information-theoretic surrogate measure.

\begin{definition}[Conditional Mutual Information]
\label{def:cmi}
The \textbf{Conditional Mutual Information} (CMI) between two random variables $A$ and $B$ given a third random variable $C$ measures the expected amount of information that $A$ and $B$ share, conditioned on $C$. It is defined as the expected Kullback-Leibler (KL) divergence between the conditional joint distribution and the product of the conditional marginal distributions:
\vspace{-0.5em}
{\small
\begin{equation}
\label{eq:cmi_def}
\begin{aligned}
&I(A; B\! \mid \!C) \!=\! \\ &\mathbb{E}_{p(C)} \! \left[ D_{\mathrm{KL}}\!\Bigl(p(A, B \!\mid\! C=c)\! \,\big\|\, \!p(A \!\mid\! C=c)p(B \!\mid\! C=c)\Bigr) \right].
\end{aligned}
\end{equation}
}
\vspace{-0.5em}
This can be expressed over the entire distribution of $C$ as:
{\small
\begin{equation}
\label{eq:cmi_integral_def}
I(A; B \!\mid\! C) \!=\! \!\int\! D_{\mathrm{KL}}\!\Bigl(p(A, B \!\mid\! C)\! \,\big\|\, \!p(A \!\mid\! C)p(B \!\mid\! C)\!\Bigr) d{p(C)}.
\end{equation}
}
\end{definition}
\vspace{-0.5em}

CMI provides a principled way to measure conditional dependence due to its fundamental properties.

\begin{lemma}[Properties of CMI]
\label{lemma:cmi_properties}
Conditional mutual information is non-negative, i.e., $I(A; B \mid C) \geq 0$. Furthermore, $I(A; B \mid C) = 0$ if and only if $A$ and $B$ are conditionally independent given $C$ ($A \perp B \mid C$). See \textit{Appendix 1} for a detailed proof.
\end{lemma}

\begin{lemma}[Symmetry of CMI]
\label{lemma:cmi_symmetry}
Conditional mutual information is symmetric in its primary arguments:
\begin{equation}
\label{eq:cmi_symmetry}
I(A; B \mid C) = I(B; A \mid C).
\end{equation}
\end{lemma}

\subsection{Variational Autoencoder}
\label{sec:prelim_vae}

A Variational Autoencoder (VAE)~\cite{kingma2022autoencodingvariationalbayes} is a generative model that learns a latent representation of data. It uses an \textbf{encoder} network, $q_{\phi}(z \mid x)$, to map an input sample $x$ to a latent distribution, and a \textbf{decoder} network, $p_{\theta}(x \mid z)$, to reconstruct the input from a latent sample $z$.

The model is trained by minimizing the negative Evidence Lower Bound (ELBO):
\vspace{-0.5em}
\begin{equation}
\begin{aligned}
\label{vae_loss}
&\mathcal{L}_{\text{VAE}}(\theta, \phi)
\;=\; \\
&\underbrace{\mathbb{E}_{q_{\phi}(z \mid x)}
\bigl[-\log p_{\theta}(x \mid z)\bigr]}_{\text{Reconstruction Loss}}
\;+\;
\underbrace{\mathrm{KL}\Bigl(q_{\phi}(z \mid x)\,\big\|\,p_{\theta}(z)\Bigr)}_{\text{KL Divergence}},
\end{aligned}
\end{equation}
where the first term measures reconstruction accuracy and the second term is a regularizer that pushes the learned latent distribution $q_{\phi}(z \mid x)$ towards a prior $p_{\theta}(z)$, which is typically a standard Gaussian $\mathcal{N}(0, I)$.

\subsection{Total Correlation Loss}

To enforce statistical independence among latent variables, \textbf{FactorVAE} \cite{kim2018disentangling} introduces a penalty on the \textbf{Total Correlation (TC)}. TC is the Kullback-Leibler (KL) divergence between the aggregate posterior, $q(z)$, and the product of its marginals, $\prod_j q(z_j)$:
\vspace{-0.5em}
{\small
\begin{equation}
\label{eq:TC loss}
    L_{TC} = \text{KL}\left(q(z) \middle\| \prod_j q(z_j)\right)
\end{equation}
}
\vspace{-0.5em}

As this term is intractable to compute directly, it is approximated using a discriminator, $D$, which is trained to distinguish between samples from $q(z)$ and samples from the product of marginals (approximated by permuting dimensions across a batch). The encoder, in turn, is trained to minimize the following adversarial loss, thereby fooling the discriminator and reducing the TC:
{\small
\begin{equation}
    L_{TC} \approx \mathbb{E}_{q(z)} \left[ \log \frac{D(z)}{1 - D(z)} \right]
\end{equation}
}

\section{Method}
\label{sec:method}

We first present the problem definition, model components, and architecture, which serve as crucial foundations for the subsequent sections. 

\subsection{CAD-VAE} \label{sec3.1}

Let $\mathcal{D} = \{(x_i, y_i, s_i)\}_{i=1}^N$ denote a dataset consisting of triplets, where $x_i$ denotes an input sample (e.g., an image), $y_i$ is the label of $x_i$ corresponding to target property $Y$, and $s_i$ is the label corresponding to the sensitive property $S$. The value range of \(Y\) and \(S\) is \(\mathcal{Y}\) and \(\mathcal{S}\), respectively, i.e., \(y \in \mathcal{Y}\) and \(s \in \mathcal{S}\). 

As discussed in Introduction, correlated information between the target attribute \( Y \) and the sensitive attribute \( S \) is pervasive in disentanglement learning. To address this, we introduce an additional latent code \( z_R \) to explicitly model this correlated information. 

The goal is to learn a latent representation that factorizes the information relevant to $Y$, the information relevant to $S$, the shared information between $Y$ and $S$, and the background or irrelevant factors. 
As defined below:
\begin{itemize}
    \item $z_X$: captures task-irrelevant information.
    \item $z_Y$: encodes the information strongly correlated with $Y$.
    \item $z_S$: encodes the information strongly correlated with $S$.
    \item $z_R$: represents the \emph{shared}  information between $Y$ and $S$.
\end{itemize}
Hence, for a single observation, the corresponding latent variable set is $z := (z_X, z_Y, z_S, z_R)$. 

\begin{figure}[tp]
    \centering
    \includegraphics[width=0.85\linewidth]{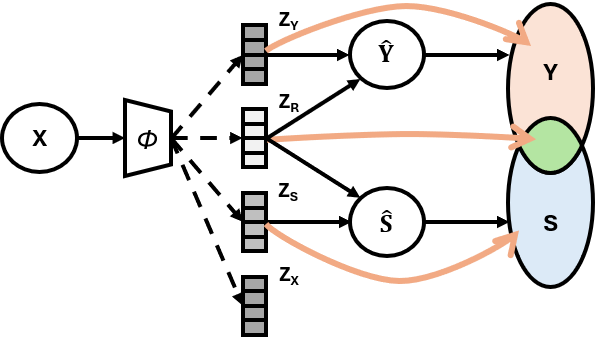}
    \caption{\textbf{Illustration of the data flow.} The \textcolor{orange}{orange lines} connect the information in the observed space and their corresponding latent codes.}
    \label{fig:casual}
    \vspace{-1.5em}
\end{figure}

The latent code \( z_R \) isolates the overlapping information between \( Y \) and \( S \), which allows the primary latent codes \( z_Y \) and \( z_S \) to remain free of unwanted correlations while preserving the model's predictive power \cite{creager2019flexibly, kim2021counterfactual, Jang_2024_CVPR}. From a causal perspective as illustrated in Figure~\ref{fig:casual}, if \( Y \) and \( S \) are conditionally independent given \( z_R \), then \( z_R \) acts as their common cause, thereby promoting the independence of \( z_Y \) and \( z_S \). 

To learn such latent code, we employ the Variational Autoencoder (VAE) framework as our backbone. The model is trained to minimize the negative ELBO, $\mathcal{L}_{\text{VAE}}$, as defined in Eq.~\eqref{vae_loss}.

 In addition, we introduce four classifiers to enforce different constraints, including:
 \begin{itemize}
     \item Enforcing $z_Y$ and $z_S$ to capture sufficient information ensures that attributes $Y$ and $S$ can be recovered correspondingly in alignment with $z_R$.
     \item Eliminating information leakage (see in subsequent section)
     \item Encouraging $z_R$ encapsulate only the correlated information $Y \cap S$ (see in subsequent section)
 \end{itemize}
 Here, we first present the training method and loss function of each classifier.
\begin{itemize}
    \item $f_y(z_Y, z_R)$ is a classifier that predicts $\hat{y}$ from $(z_Y, z_R)$;
    \item $f_s(z_S, z_R)$ is a classifier that predicts $\hat{s}$ from $(z_S, z_R)$;
    \item $f_{y\_op}(z_S)$ is an \emph{opponent} classifier that attempts to predict $\hat{y}$ from $z_S$;
    \item $f_{s\_op}(z_Y)$ is an \emph{opponent} classifier that attempts to predict $\hat{s}$ from $z_Y$.
\end{itemize}
Let $\omega_{y}$, $\omega_{s}$, $\omega_{y\_op}$, and $\omega_{s\_op}$ denote the parameters of these four classifiers, respectively. 

We define
\begin{equation}
\label{eq:cls-y}
\min_{\phi,\omega_{y}} 
\Bigl[
\mathcal{L}_y(\omega_{y}, \phi) \Bigr]
\!=\!
\mathbb{E}_{(x,y)\sim \mathcal{D}}
\Bigl[
    -\log f_y\bigl(\hat{y} \,\big\vert\, z_Y, z_R\bigr)
\Bigr]
,
\end{equation}
where $z_Y$ and $z_R$ are sampled from the encoder $q_\phi(z \mid x)$: $(z_Y, z_R) \sim q_{\phi}(z \mid x)$. 
The parameters $\omega_y$ and the encoder parameters $\phi$ are jointly updated to reduce the cross-entropy in~\eqref{eq:cls-y}, ensuring that $(z_Y, z_R)$ carry sufficient information about $Y$. Similarly, the classifier $f_s(z_S, z_R)$ predicts $s$:
\begin{equation}
\label{eq:cls-s}
\min_{\phi,\omega_{s}} 
\Bigl[
\mathcal{L}_s(\omega_{s}, \phi) \Bigr]
\!=\!
\mathbb{E}_{(x,s)\sim \mathcal{D}}
\Bigl[
    -\log f_s\bigl(\hat{s} \,\big\vert\, z_S, z_R\bigr)
\Bigr]
.
\end{equation}

To measure the information leakage, 
we introduce:
\begin{equation}
\label{eq:cls-y-op}
\min_{\omega_{y\_op}} 
\Bigl[
\mathcal{L}_{y\_op}(\omega_{y\_op}; \phi) \Bigr]
\!=\!
\mathbb{E}_{(x,y)\sim \mathcal{D}}
\Bigl[
    -\log f_{y\_op}\bigl(\hat{y} \,\big\vert\, z_S\bigr)
\Bigr]
,
\end{equation}
where $z_S \sim q_{\phi}(z \mid x)$ is produced by the frozen encoder i.e $\phi$ is \emph{not} updated during the minimization of~\eqref{eq:cls-y-op}; this network is trained to detect any $Y$-relevant information that may unintentionally exist in $z_S$. Analogously, the classifier $f_{s\_op}(z_Y)$ aims to predict $s$ given $z_Y$:
\begin{equation}
\label{eq:cls-s-op}
\min_{\omega_{s\_op}} 
\Bigl[
\mathcal{L}_{s\_op}(\omega_{s\_op}; \phi) \Bigr]
\!=\!
\mathbb{E}_{(x,s)\sim \mathcal{D}}
\Bigl[
    -\log f_{s\_op}\bigl(\hat{s} \,\big\vert\, z_Y\bigr)
\Bigr]
.
\end{equation}
Likewise, $\phi$ is fixed, and only $\omega_{s\_op}$ is updated when minimizing \eqref{eq:cls-s-op}. 

To achieve correlation-aware disentanglement learning, we propose directly minimizing the conditional mutual information between $z_Y$ and $z_S$ with respect to their corresponding opposite attributes $S$ and $Y$, conditioned on $z_R$. This approach is complemented by an explicit relevance learning strategy that constrains $z_R$ to effectively capture shared information between $Y$ and $S$ while avoiding redundant information. Detailed explanations of these strategies are provided in following section.

\subsection{Conditional Independence for Disentanglement}



Our fairness objective is to achieve independence between \( z_Y \) and \( z_S \) by enforcing conditional independence between their respective predictions, \( \hat{Y} \) and \( \hat{S} \), given the shared latent code \( z_R \). The predictions are generated by classifiers:
$\hat{Y} = f_y(z_Y, z_R)$ and $\hat{S} = f_s(z_S, z_R)$.
This objective is formally expressed as $\hat{Y} \perp \hat{S} \mid z_R$.
Following \textbf{Proposition \ref{prop:ci_factorization}}, this conditional independence is equivalent to the factorization of the conditional joint distribution:
\begin{equation}
\label{eq:CI_definition}
p_\theta(\hat{Y}, \hat{S} \mid z_R) = p_\theta(\hat{Y} \mid z_R)p_\theta(\hat{S} \mid z_R).
\end{equation}

As noted in the Preliminary section, directly minimizing the divergence between the distributions in Eq.~\eqref{eq:CI_definition} is generally intractable. We therefore adopt an information-theoretic approach and minimize the Conditional Mutual Information (CMI), $I_\phi(\hat{Y}; \hat{S} \mid z_R)$, as a tractable surrogate objective.

From \textbf{Definition \ref{def:cmi}}, the CMI is the expected KL divergence over \(z_R\):
{\small
\begin{equation}
\label{eq:CMI_kl}
\begin{aligned}
&I_\phi(\hat{Y}; \hat{S} \!\mid\! z_R) \!=\! \\ &\!\int\! \!D_{\mathrm{KL}}\!\Bigl(p_\theta(\hat{Y}, \hat{S} \!\mid\! z_R)\! \,\big\|\,\! p_\theta(\hat{Y} \!\mid\! z_R)\! \, p_\theta(\hat{S} \!\mid\! z_R)\!\Bigr) dP_{z_R}.
\end{aligned}
\end{equation}
}
According to \textbf{Lemma \ref{lemma:cmi_properties}}, minimizing $I_\phi(\hat{Y}; \hat{S} \mid z_R)$ to zero is equivalent to enforcing the conditional independence defined in Eq.~\eqref{eq:CI_definition}. Furthermore, leveraging the symmetry property from \textbf{Lemma \ref{lemma:cmi_symmetry}}, we note that $I_\phi(\hat{Y}; \hat{S} \mid z_R) = I_\phi(\hat{S}; \hat{Y} \mid z_R)$. Thus, minimizing this CMI ensures that any undesired dependence between \(\hat{Y}\) and \(\hat{S}\) not explained by \(z_R\) is removed.

\subsection{Direct Minimization of Conditional Mutual Information} \label{sec3.3}
While FADES \cite{Jang_2024_CVPR} minimizing CMI~\eqref{eq:CMI_kl} through approximation $I_\phi(\hat{Y}; S \mid z_R)$: 
$$\min_\phi 
\Bigl[I_\phi(\hat{Y}; S \mid z_R)\Bigr] = \min_\phi 
\Bigl[H_\phi(\hat{Y} \mid z_R) - H_\phi(\hat{Y} \mid S, z_R)\Bigr]$$
by reducing CMI through ground truth-based sample grouping, its reliance on batch-level sampling introduces instability. 

In contrast, our method directly minimizes CMI~\eqref{eq:CMI_kl} via a principled information-theoretic approach, providing a more robust and stable disentanglement process by avoiding sampling variance and reducing dependency on batch-specific dynamics.
To achieve CI as \eqref{eq:CI_definition}, we propose directly minimizing:
\begin{equation}
\label{CMI_additive}
\min_{\phi} 
\Bigl[
I_\phi(\hat{Y}; \hat{S} \mid z_R) + I_\phi(\hat{S}; \hat{Y} \mid z_R)
\Bigr], 
\end{equation}
where:
\begin{equation}
\label{CMI_ys_zr}
I_\phi(\hat{Y}; \hat{S} \mid z_R) = H_\phi(\hat{Y} \mid z_R) - H_\phi(\hat{Y} \mid \hat{S}, z_R), 
\end{equation}
$H_\phi(* \mid *)$ stands for the conditional entropy.
Incorporating $\hat{S} \!=\! f_s(z_S, z_R)$,
we have:
{\small
\begin{equation}
\label{CMI_ys_zr_simplify}
\begin{aligned}
    I_\phi(\hat{Y}; \hat{S} \mid z_R) &= H_\phi(\hat{Y} \mid z_R) - H_\phi(\hat{Y} \mid \hat{S}, z_R) \\
    &= H_\phi(\hat{Y} \mid z_R) - H_\phi(\hat{Y} \mid f_s(z_S, z_R), z_R) \\
    &= H_\phi(\hat{Y} \mid z_R) - H_\phi(\hat{Y} \mid z_S, z_R) \\
    &= I_\phi(\hat{Y}; z_S \mid z_R),
\end{aligned}
\end{equation}
}
from the same transformation (see detailed derivation in \textit{Appendix 2}):
\begin{align}
\label{CMI_sy_zr_simplify}
I_\phi(\hat{S}; \hat{Y} \mid z_R) &= I_\phi(\hat{S}; z_Y \mid z_R)\nonumber \\
&= H_\phi(\hat{S} \mid z_R) - H_\phi(\hat{S} \mid z_Y, z_R).
\end{align}
Therefore, with the introduction of the correlated latent code $z_R$ that captures all relevant information between $\hat{Y}$ and $\hat{S}$:
\begin{equation}
\begin{aligned}
\min_{\phi} 
\Bigl[
    &I_\phi(\hat{Y}; \hat{S} \mid z_R) + I_\phi(\hat{S}; \hat{Y} \mid z_R)
\bigr] \\
    &\equiv 
\min_{\phi} 
\Bigl[
    I_\phi(\hat{Y}; z_S \mid z_R) + I_\phi(\hat{S}; z_Y \mid z_R)
\Bigr].
\end{aligned}
\end{equation}

For the minimization of $I_\phi(\hat{Y}; z_S \mid z_R)$, as shown in \eqref{CMI_ys_zr_simplify} 
, since $z_R$ is given as a condition, we can consider this CMI formula as a function where the independent variable is $z_S$ and the dependent variable is $\hat{Y}$, as shown as:
\begin{equation}
\label{CMI_func_y_zs}
\mathcal{L}_{\hat{Y}}(z_S) = H_\phi(\hat{Y} \mid z_R) - H_\phi(\hat{Y} \mid z_S, z_R), 
\end{equation}
where $z_R$ is determined here, $H_\phi(\hat{Y} \mid z_R)$ is a constant, henceforth we need to minimize $-H_\phi(\hat{Y} \mid z_S, z_R)$. Empirically, we directly minimize the lower bound of it: $-H_\phi(\hat{Y} \mid z_S)$, since: $-H_\phi(\hat{Y} \mid z_S, z_R) \geq -H_\phi(\hat{Y} \mid z_S)$.
Symmetrically, the minimization of $I_\phi(\hat{S}; z_Y \mid z_R)$ shown in \eqref{CMI_sy_zr_simplify} is the same concept, see \textit{Appendix 2} for detailed derivation. In this optimization process, $z_R$ is responsible for containing any correlation information between target attribute $Y$ and sensitive attribute $S$. 

After these simplifications, we introduce the CMI loss to minimize \eqref{CMI_additive}:
{\small
\begin{equation}
\label{CMI_loss}
\min_{\phi} 
\Bigl[
\mathcal{L}_{\text{CMI}}(\omega_{y\_op}; \omega_{s\_op}; \phi) \Bigr] \!
 \!=\! - (H_\phi(\hat{Y} \mid z_S) + H_\phi(\hat{S} \mid z_Y)),
\end{equation}
}
where only update encoder parameters $\phi$. Utilizing opponent classifier $f_{y\_op}(z_S)$, the entropy term calculation are shown as below:
 {\small
    \begin{equation}
    \begin{aligned}
    \label{eq:H-Y-zS}
    H_{\phi}(\hat{Y}\mid z_S)
    &\!=\!
    \mathbb{E}_{q_{\phi}(z_S \mid x)}\!
    \biggl[
      -\!\sum_{\,\hat{y} \in \mathcal{Y}}\,\!p_{\theta}(\hat{y} \mid z_S)\,\!\log \,p_{\theta}(\hat{y} \mid z_S)
    \biggr] \\
    &\!=\!
    \frac{1}{\lvert B\rvert}\,\!\sum_{i=1}^{\lvert B\rvert}\,\!\sum_{\,\hat{y} \in \mathcal{Y}}\,\!
    \biggl[
      -\!
      p_{\theta}\bigl(\hat{y} \mid z_S^{(i)}\bigr)\!
      \log p_{\theta}\bigl(\hat{y} \mid z_S^{(i)}\bigr)
    \biggr],
    \end{aligned}
    \end{equation}
    }
    where 
    $
    p_{\theta}\bigl(\hat{y} \mid z_S^{(i)}\bigr)
    \!=\!
    f_{y\_op}\bigl(z_S^{(i)}\bigr).
    $
Here, $z_S^{(i)}$ denotes the $z_S$ sample from the $i$-th element in a mini-batch of size $|B|$; the distribution $q_{\phi}(z_S \mid x)$ is given by the encoder. Similar calculation to $H_\phi(\hat{S} \mid z_Y)$, see \textit{Appendix 3} for completed calculation formula. During this optimization process, the opponent classifier parameters $\omega_{y\_op}$ and $ \omega_{s\_op}$ are frozen.

\begin{table*}[htbp]
\centering
\caption{\textbf{Evaluation of downstream classification tasks on various datasets from learned representation.} Best in \textbf{bold}, second in \textcolor{red}{red}.}
\vspace{-1em}
\label{tab_fair_cls}
\renewcommand{\arraystretch}{1} 
\setlength{\tabcolsep}{3pt} 
\begin{adjustbox}{width=\textwidth,center}
\begin{tabular}{l||ccc|ccc|ccc|ccc} 
\Xhline{2\arrayrulewidth}
\rowcolor{gray!20} \multicolumn{13}{c}{\textbf{Downstream Classification Performance}} \\  
\rowcolor{gray!20}  
\textbf{Methods} & \multicolumn{3}{c|}{CelebA~\cite{liu2015deep}} & \multicolumn{3}{c|}{UTKFace~\cite{zhang2017age}} & \multicolumn{3}{c|}{Dogs and Cats~\cite{parkhi2012cats}} & \multicolumn{3}{c}{Color bias MNIST~\cite{Kim_2019_CVPR}} \\  
\rowcolor{gray!20}  
& Acc $\uparrow$ & EOD $\downarrow$ & DP $\downarrow$ & Acc $\uparrow$ & EOD $\downarrow$ & DP $\downarrow$ & Acc $\uparrow$ & EOD $\downarrow$ & DP $\downarrow$ & Acc $\uparrow$ & EOD $\downarrow$ & DP $\downarrow$ \\ 
\Xhline{1.5\arrayrulewidth} 

FADES~\cite{Jang_2024_CVPR} \scriptsize{[CVPR'24]} & 0.918 & \textcolor{red}{0.034} & 0.135 & 0.812 & \textcolor{red}{0.059} & 0.139 & \textcolor{red}{0.769} & 0.058 & 0.086 & \textcolor{red}{0.973} & 0.094 & 0.160 \\ 
GVAE~\cite{ding2020guided} \scriptsize{[CVPR'20]} & \textcolor{red}{0.919} & 0.047 & 0.131 & \textcolor{red}{0.819} & 0.204 & 0.197 & 0.748 & 0.064 & 0.131 & 0.961 & 0.109 & 0.176 \\ 
FFVAE~\cite{creager2019flexibly} \scriptsize{[PMLR'19]} & 0.892 & 0.076 & 0.072 & 0.766 & 0.269 & 0.201 & 0.729 & 0.059 & 0.110 & 0.952 & \textcolor{red}{0.081} & \textbf{0.092} \\ 
ODVAE~\cite{sarhan2020fairness} \scriptsize{[ECCV'20]} & 0.886 & 0.039 & 0.103 & 0.736 & 0.165 & 0.210 & 0.689 & 0.051 & \textbf{0.038} & 0.957 & 0.247 & 0.162 \\ 
FairDisCo~\cite{liu2022fair} \scriptsize{[KDD'22]} & 0.839 & 0.074 & \textbf{0.051} & 0.766 & 0.266 & 0.200 & 0.680 & 0.115 & 0.111 & 0.949 & 0.129 & 0.136 \\ 
FairFactorVAE~\cite{liu2023fair} & 0.914 & 0.055 & 0.136 & 0.720 & 0.096 & \textbf{0.134} & 0.707 & \textcolor{red}{0.055} & 0.110 & 0.957 & 0.096 & 0.128 \\ 
\rowcolor{gray!15} \textbf{CAD-VAE (Ours)} & \textbf{0.939} & \textbf{0.021} & \textcolor{red}{0.065} & \textbf{0.828} & \textbf{0.045} & \textcolor{red}{0.137} & \textbf{0.781} & \textbf{0.048} & \textcolor{red}{0.069} & \textbf{0.984} & \textbf{0.076} & \textcolor{red}{0.108} \\ 
\Xhline{2\arrayrulewidth}
\end{tabular}
\end{adjustbox}
\vspace{-1.5em}
\end{table*} \label{sec4.1}

\subsection{Learning Relevance Between Target And Sensitive Information} \label{sec3.4}
To encourage $z_R$ capture and only capture the shared information relevant to target property and sensitive property, as well as $z_Y$, $z_S$ capture main information of $Y$ and $S$ attributes respectively, we propose to maximize the conditional mutual information as Learning Relevance Information loss:
 {\small
\begin{equation}
\label{LRI_loss}
\min_{\phi} 
\Bigl[
\mathcal{L}_{\text{LRI}}(\omega_{y}; \omega_{s}; \phi) \Bigr] \!=\! -(I_\phi(\hat{Y}; Y \mid z_R) + I_\phi(\hat{S}; S \mid z_R)), 
\end{equation}
}
where:
\begin{equation}
\label{CRI_yy_zr}
I_\phi(\hat{Y}; Y \mid z_R) = H_\phi(\hat{Y} \mid z_R) - H_\phi(\hat{Y} \mid Y, z_R), 
\end{equation}
\begin{equation}
\label{CRI_ss_zr}
I_\phi(\hat{S}; S \mid z_R) = H_\phi(\hat{S} \mid z_R) - H_\phi(\hat{S} \mid S, z_R), 
\end{equation} 
$H_\phi(* \mid *)$ stands for the conditional entropy. Maximizing $H_\phi(\hat{Y} \mid z_R)$, $H_\phi(\hat{S} \mid z_R)$ avoid $z_R$ capture all information of $Y$ or $S$ solely, which will lead to the Information Bottleneck phenomenon\cite{Jang_2024_CVPR, creager2019flexibly, kim2018disentangling} i.e $z_R$ capture all the information about target attribute $Y$ and sensitive attribute $S$: $Y \cup S$, degenerating disentanglement performance. On the other hand, minimizing $H_\phi(\hat{Y} \mid Y, z_R)$ and $H_\phi(\hat{S} \mid S, z_R)$ enforce $z_R$ determine $\hat{Y}$ or $\hat{S}$ only within each $Y$ or $S$ subgroup, so that encourage $z_R$ capture information both relevant to $Y$ and $S$: $Y \cap S$. 

In contrast to FADES\cite{Jang_2024_CVPR}, which exhibits a conflict between the disentanglement term and the regularization term, our method achieves orthogonality between CMI~\eqref{CMI_loss} and LRI~\eqref{LRI_loss}, ensuring fair disentanglement while preserving robust representations.
See Appendix 5 for details.


For entropy calculation of $H_\phi(\hat{Y} \mid z_R)$, We approximate $p_\theta(\hat{y} \mid z_R)$ by marginalizing over $z_Y$:
   \begin{equation}
   \begin{aligned}
   \label{cal_P_y_zr}
   p_\theta(\hat{y} \mid z_R^{(k)}) &= \mathbb{E}_{p(x)} \left[ \mathbb{E}_{q_\phi(z_Y \mid x)} \left[ p_\theta(\hat{y} \mid z_Y, z_R^{(k)}) \right] \right] \\ &\approx \frac{1}{|B|}\sum_{i=1}^{|B|} p_\theta(\hat{y} \mid z_Y^{(i)}, z_R^{(k)}), 
   \end{aligned}
   \end{equation}
   where 
   $
    p_\theta(\hat{y} \mid z_Y^{(i)}, z_R^{(k)})
    \!=\!
    f_y\bigl(z_Y^{(i)}, z_R^{(k)}\bigr)
    $, 
   then
    {\small
   \begin{equation}
   \begin{aligned}
   \label{cal_H_y_zr}
   H_\phi(\hat{Y} \mid z_R) \!=\! \mathbb{E}_{q_\phi(z_R \mid x)}\left[-\sum_{\hat{y} \in \mathcal{Y}} p_\theta(\hat{y} \mid z_R)\log p_\theta(\hat{y} \mid z_R)\right] \\
    \!=\! \frac{1}{|B|} \sum_{i=1}^{|B|} \sum_{\hat{y} \in \mathcal{Y}} \left[ -p_\theta(\hat{y} \mid z_R^{(i)}) \log p_\theta(\hat{y} \mid z_R^{(i)}) \right].
    \end{aligned}
   \end{equation}
   }

As for the calculation of conditional entropy term $H_\phi(\hat{Y} \mid Y, z_R)$, we regard known condition $Y$ as attribute to grouping samples in a mini-batch of size $|B|$, and calculate the entropy term by marginalizing over $z_Y$ within each group, $p_{\theta}(\hat{y} \mid z_R, y)$ can be computed for $z_R^{(k)}$ sampled from an instance \( x^{(k)} \in B_y \) as:
 {\small
\begin{equation}
\begin{aligned}
\label{cal_P_y_zr_y}
p_{\theta} (\hat{y} | z_R^{(k)}, y) &= \mathbb{E}_{p(x\mid Y=y)} \left[ \mathbb{E}_{q_{\phi}(z_Y \mid x)} \left[ p_{\theta} (\hat{y} \mid z_Y, z_R^{(k)}) \right] \right] \\
&\approx \frac{1}{|B_y|} \sum_{i=1}^{|B_y|} p_{\theta} (\hat{y} | z_Y^{(i)}, z_R^{(k)}),
\end{aligned}
\end{equation}
}
where \( B_y \) denotes a subset of the batch with \( Y = y \). Then the conditional entropy can be computed as: 
 \vspace{-0.5em}
 {\small
\begin{equation}
\begin{aligned}
\label{cal_H_Y_Y_zr}
&H_{\phi} (\hat{Y} \mid Y, z_R)= \\ &\mathbb{E}_{(x,y)\sim \mathcal{D}}\!\left[ \!\mathbb{E}_{q_\phi(z_R \mid x)}\!\left[ \!-\!\sum_{\hat{y} \in \mathcal{Y}} \!p_\theta(\hat{y} \!\mid\! z_R, y)\!\log p_\theta(\hat{y}\! \mid\! z_R, \!y)\right] \! \right] \!\\
&= \frac{1}{|B|}\! \sum_{y \in \mathcal{Y}}\! \sum_{i=1}^{|B_y|}\! \sum_{\hat{y} \in \mathcal{Y}} \left[ -p_{\theta} (\hat{y} | z_R^{(i)}, y) \!\log p_{\theta} (\hat{y} | z_R^{(i)}, y) \right].
\end{aligned}
\end{equation}
}
The calculation of $H_\phi(\hat{S} \mid z_R)$ and $H_\phi(\hat{S} \mid S, z_R)$ are similar to $H_\phi(\hat{Y} \mid z_R)$, $H_{\phi} (\hat{Y} \mid Y, z_R)$ respectively, see \textit{Appendix 3} for completed calculation formula. Plugging these estimates\eqref{cal_H_y_zr}\eqref{cal_H_Y_Y_zr} back into \eqref{LRI_loss}, shared feature between $Y$ and $S$ will be learned in $z_R$ while getting rid of Information Bottleneck phenomenon.
Note that the classifier parameters $\omega_y$ and $\omega_s$ remain frozen when optimizing \eqref{LRI_loss}.

\subsection{Final Objective Function}
To integrate the above components into a coherent training framework, we employ the two-step optimization strategy defined in~\eqref{eq:final-objective-tc} and~\eqref{eq:final-opponent}. 
 \vspace{-0.7em}
\begin{equation}
\label{eq:final-objective-tc}
\begin{aligned}
&\min_{\theta,\phi,\omega_{y},\omega_{s}}
\Bigl[
    \mathcal{L}_{\text{VAE}}(\theta,\phi)
    + \bigl(\mathcal{L}_y(\omega_{y},\phi)
    + \mathcal{L}_s(\omega_{s},\phi)\bigr)
\Bigr] \!
\!+\! 
\min_{\phi} 
\Bigl[ \\
    &\lambda_{\text{CMI}}\mathcal{L}_{\text{CMI}}(\omega_{y\_op},\omega_{s\_op};\phi)
    \!+\! \mathcal{L}_{\text{TC}}(\phi)
\!+\! 
    \lambda_{\text{LRI}}\,\mathcal{L}_{\text{LRI}}(\omega_{y},\omega_{s};\phi)
\Bigr] 
\end{aligned}
\end{equation}

Specifically, in~\eqref{eq:final-objective-tc}, we jointly update $(\theta,\phi,\omega_{y},\omega_{s})$ by minimizing the VAE loss~\eqref{vae_loss} alongside the main classification losses~\eqref{eq:cls-y} and~\eqref{eq:cls-s}, which together reformulate the ELBO. We further include the CMI loss~\eqref{CMI_loss} to reduce unwanted information leakage, the LRI loss~\eqref{LRI_loss} to capture shared patterns in $z_R$, and the TC penalty~\eqref{eq:TC loss} to promote factorization among the latent codes $(z_Y, z_R, z_S)$. 
 \vspace{-0.5em}
\begin{equation}
\label{eq:final-opponent}
    \min_{\omega_{y\_op},\omega_{s\_op}}
\Bigl[
    \mathcal{L}_{y\_op}(\omega_{y\_op};\phi)
    + \mathcal{L}_{s\_op}(\omega_{s\_op};\phi)
\Bigr] 
\end{equation}
 \vspace{-0.7em}

In parallel, the second procedure~\eqref{eq:final-opponent} optimizes $(\omega_{y\_op}, \omega_{s\_op})$ by minimizing the opponent classification losses~\eqref{eq:cls-y-op} and~\eqref{eq:cls-s-op} while holding $\phi$ fixed. 

The hyperparameters $\lambda_{CMI},\lambda_{LRI} > 0$ control the relative importance of these terms, ensuring each network component learns its designated function while enforcing minimal information leakage, preserving shared information in $z_R$ and maintaining the salient factors for $Y$ and $S$ in $z_Y$ and $z_S$ respectively. See hypermeter analysis in \textit{Appendix 6}.

\section{Experiment}
\label{sec:experiment}

To ensure a rigorous and comprehensive evaluation, we conduct experiments comparing our proposed method with a diverse set of state-of-the-art approaches across multiple categories of learning paradigms in various tasks. Specifically, we include FairFactorVAE~\cite{liu2023fair}, FairDisCo~\cite{liu2022fair}, FFVAE~\cite{creager2019flexibly}, GVAE~\cite{ding2020guided}, ODVAE~\cite{sarhan2020fairness} and FADES~\cite{Jang_2024_CVPR}, shown in Section 2.1. As for traditional correlation-aware learning that is discussed in Section 2.2, since it require additional annotated data to build causal graph, we except them in our experiment. 

\subsection{Fair Classification}

The objective of fair classification is to achieve a balance between minimizing fairness violations and maintaining high predictive performance. To evaluate the effectiveness of our proposed method, we conduct experiments on a diverse set of benchmark fairness datasets. For facial attribute classification tasks, we utilize the CelebA \cite{liu2015deep} and UTKFace \cite{zhang2017age} datasets. Following prior works \cite{wang2022fairness, xu2020investigating, zeng2022boosting, Jang_2024_CVPR}, we set the CelebA classification task to predict the ``Smiling" attribute, while for UTKFace, the objective is to classify whether a person depicted in the image is over 35 years old, with gender serving as the sensitive attribute. Additionally, the Dogs and Cats dataset \cite{parkhi2012cats} is used to distinguish between dogs and cats, with fur color as the sensitive attribute. Furthermore, we assess fair classification performance using the Colored MNIST dataset \cite{kim2021biaswap, Kim_2019_CVPR, nam2020learning}, which incorporates a controlled color bias in the standard MNIST dataset to simulate spurious correlations. To assess fairness violations, we use standard metrics including Demographic Parity (DP) \cite{barocas2016big} and Equalized Odds (EOD) \cite{hardt2016equality}. See detailed experimental setup in \textit{Appendix 6}. The result of fair classification can be seen in Table~\ref{tab_fair_cls}. Across all evaluated datasets, our method consistently achieves state-of-the-art classification accuracy and fairness, validating its effectiveness in robust disentanglement by preserving high-quality target-related information while minimizing sensitive attribute leakage.

\begin{table}[t]
\centering
\renewcommand{\arraystretch}{1.0}
\setlength{\tabcolsep}{4pt}
\caption{\textbf{Fair Classification on 95\% Color Bias MNIST}}
\label{tb:fair_classification_imbalance}
\vspace{-1em}
\begin{adjustbox}{width=\columnwidth}
\begin{tabular}{l|cccccc}
\Xhline{1pt}
\rowcolor{gray!20}
Metric & Ours & FADES & GVAE & FFVAE & ODVAE & FairFactorVAE \\
\hline
Accuracy $\uparrow$ & \textbf{0.867} & 0.782 & 0.771 & 0.744 & 0.721 & \textcolor{red}{0.807} \\
EOD $\downarrow$    & \textbf{0.141} & \textcolor{red}{0.174} & 0.244 & 0.190 & 0.210 & 0.195 \\
DP $\downarrow$     & \textbf{0.167} & \textcolor{red}{0.201} & 0.265 & 0.213 & 0.262 & 0.221 \\
\Xhline{1pt}
\end{tabular}
\end{adjustbox}
\vspace{-1em}
\end{table}
 To assess the robustness and effectiveness of these methods, we further conduct classification experiments under an extreme imbalance bias setting, which is commonly encountered in practical applications. In the MNIST experiment, we set the color bias rate to 95\% to simulate a strong correlation between the target attribute and the sensitive attribute. The results, shown in Table~\ref{tb:fair_classification_imbalance}, demonstrate that our method outperforms existing approaches in both disentanglement ability and robust representation preservation.

\begin{figure}[tp]
    \centering
    \includegraphics[width=1\linewidth]{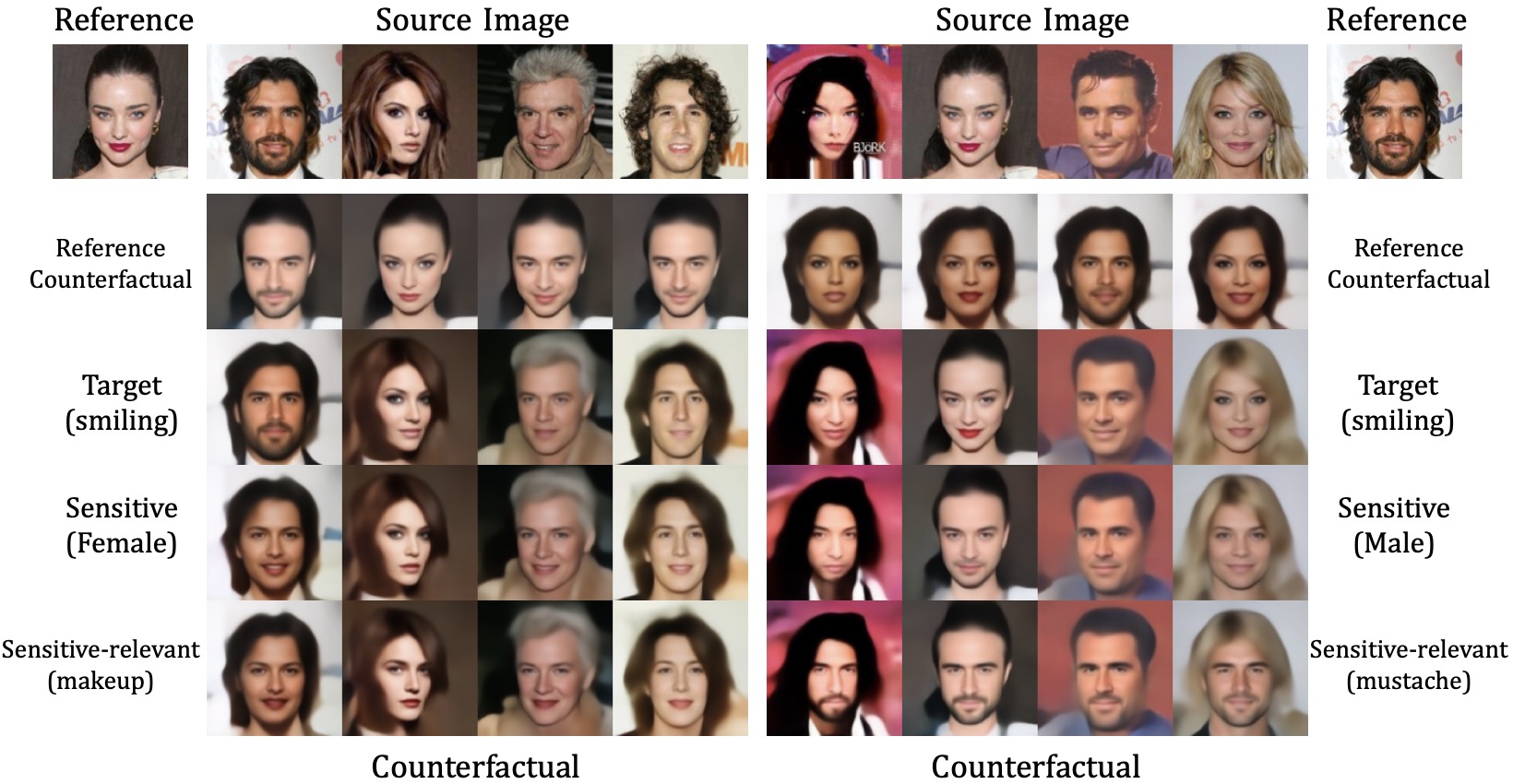}
    \vspace{-2em}
    \caption{\textbf{Examples of Fair Counterfactual Generation}. Zoom in to check. The first row shows the source and reference images. Rows 2–5 display counterfactuals obtained by replacing latent subspaces $z_X$, $z_Y$, $z_S$, and $[z_S, z_R]$, respectively. Notably, the replacement with $[z_S, z_R]$ (row 5) naturally adapts sensitive features for different sensitive attributes without domain knowledge. (mustache for men and makeup for women).}
    \label{fig:main_counterfactual_gene}
    \vspace{-1.5em}
\end{figure}

\subsection{Fair Counterfactual Generation} \label{sec4.2}

We evaluate our approach on the CelebA dataset \cite{liu2015deep}, a widely-used benchmark for facial attribute manipulation. We select \textit{Smiling} as the target label $Y$ and \textit{Gender} as the sensitive attribute $S$. 
In our experiments, we substitute specific latent code of source images with reference images, including $z_X$, $z_Y$, $z_S$, and $[z_S, z_R]$.
Figure~\ref{fig:main_counterfactual_gene} illustrates the generated counterfactuals, with the first row showing source and reference images and subsequent rows demonstrating the effects of substituting each latent code. The experimental results demonstrate the effectiveness of our method in generating fair counterfactuals. As shown in Figure~\ref{fig:main_counterfactual_gene}, substituting $[z_S, z_R]$ (Row 5) leads to a natural adaptation of sensitive-relevant features without domain-specific knowledge. For instance, the model automatically adds makeup to female images and a mustache to male images, highlighting the semantic alignment of $z_R$ with both the target and sensitive attributes. Compared to substituting only $z_S$, our approach achieves more interpretable translation, ensuring that fairness is maintained throughout the counterfactual generation. More fair counterfactual generation experiment results can be seen in \textit{Appendix 7}. 

To quantitatively assess the quality of the generated counterfactuals, we compare evaluation metrics between the direct reconstruction of the input image and the reconstructions obtained by randomly permuting $z_Y$ and $z_S$ within the evaluation set. Specifically, we use the FID~\cite{heusel2017gans, Jang_2024_CVPR} to assess reconstruction fidelity and the Inception Score (IS) \cite{chong2020effectively} to evaluate semantic and perceptual quality. Lower $\Delta \text{FID}$ values indicate minimal distortion and higher translation quality, while lower $\Delta \text{IS}$ values suggest that semantic and perceptual attributes are well preserved. Detailed experimental settings are provided in \textit{Appendix 6}.

\begin{figure*}[h]
    \centering
    \includegraphics[width=1\linewidth]{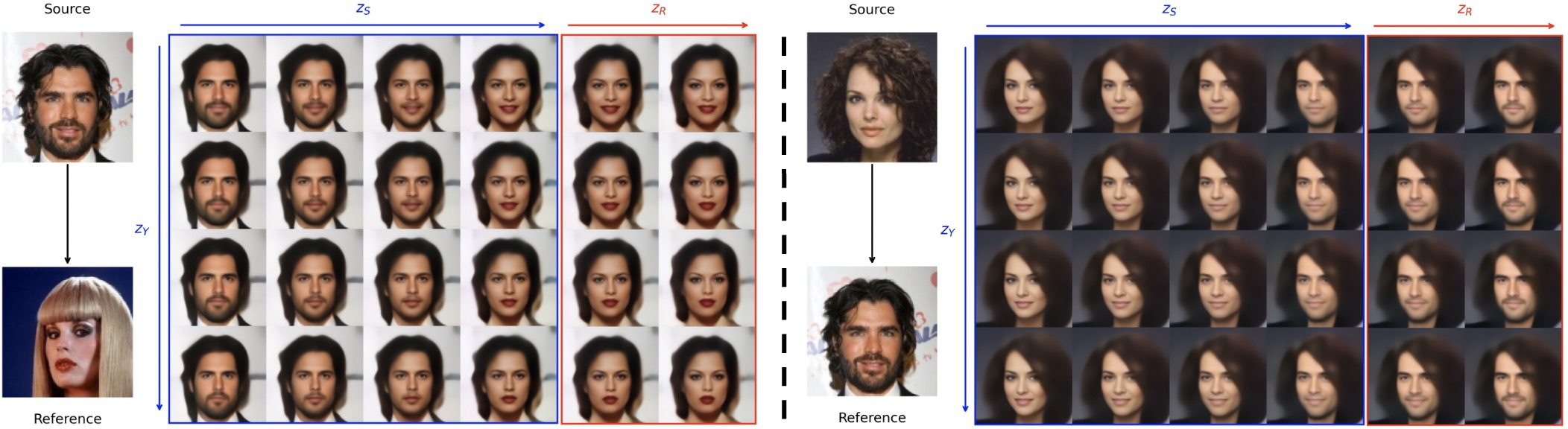}
    \includegraphics[width=1\linewidth]{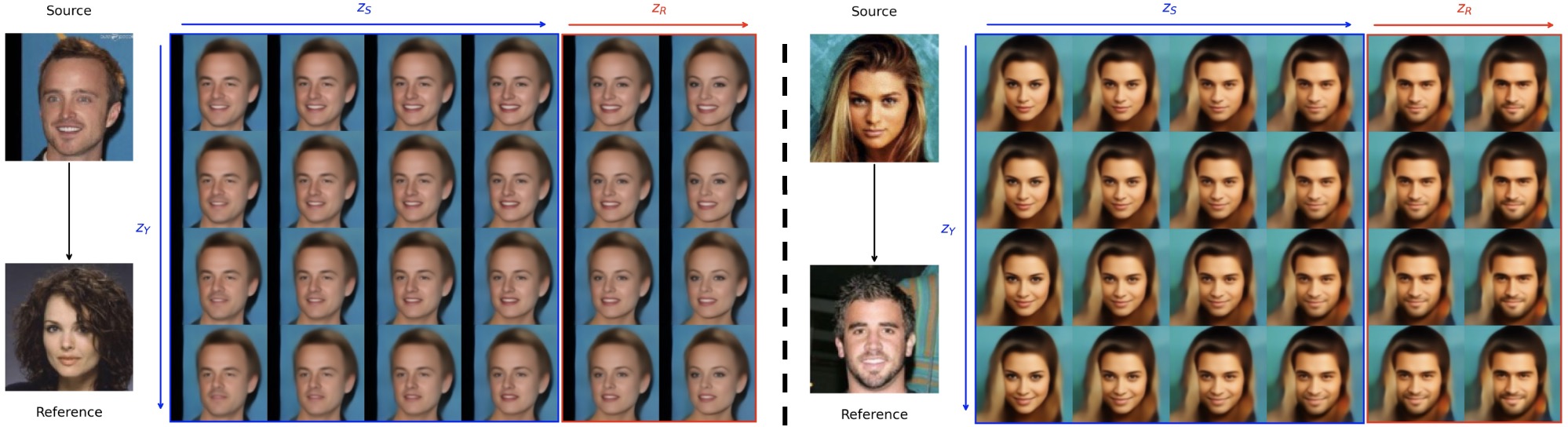}
    \vspace{-1em}
    \caption{\textbf{Examples of Fair Fine-Grained Image Editing}. Zoom in to check. The leftmost column shows the source and reference images. The blue-framed section displays images generated by interpolating \textcolor{blue}{$z_Y$} and \textcolor{blue}{$z_S$} (with \textcolor{red}{$z_R$} and $z_X$ fixed), where the horizontal axis varies \textcolor{blue}{$z_S$} and the vertical axis varies \textcolor{blue}{$z_Y$}. The red-framed section illustrates images produced by interpolating \textcolor{blue}{$z_Y$} and \textcolor{red}{$z_R$} (with \textcolor{blue}{$z_S$} fully replaced by the reference and $z_X$ constant). Modification in one latent code minimally affecting others, harness \textcolor{red}{$z_R$} to edit sensitive relevant feature(makeup or mustache).}
    \label{fig:sup_interpolation}
    \vspace{-1.5em}
\end{figure*}



\begin{table}[h]
\centering
\caption{\textbf{FID and IS difference between original reconstruction and perturbed target/sensitive codes' reconstruction.}}
\vspace{-1em}
\label{tb:quantitative_counterfactual}
\renewcommand{\arraystretch}{1} 
\setlength{\tabcolsep}{3pt} 
\begin{adjustbox}{width=\columnwidth,center}
\begin{tabular}{c||cccccc} 
\Xhline{2\arrayrulewidth}
\rowcolor{gray!20}
& \textbf{CAD-VAE} & FADES & GVAE & FFVAE & ODVAE & FairFactorVAE \\ 
\hline
$\Delta \textit{\text{FID}} \downarrow$ & \textbf{1.072} & \textcolor{red}{1.167} & 3.710 & 1.409 & 14.647 & 6.239  \\ 
\hline
$\Delta \textit{\text{IS}} \downarrow$ & \textbf{1.214} & \textcolor{red}{2.379} & 3.148 & 3.829 & 6.113 & 5.378 \\ 
\Xhline{2\arrayrulewidth}
\end{tabular}
\end{adjustbox}
\vspace{-0.5em}
\end{table}
Quantitative analysis in Table~\ref{tb:quantitative_counterfactual} further validates our approach. Our method achieves both lower $\Delta \text{FID}$ and $\Delta \text{IS}$ compared to other fair representation learning methods, demonstrating that our fair counterfactual generation approach renders counterfactuals with superior image quality and minimal distortion. 

\subsection{Fair Fine-Grained Image Editing} \label{sec4.3}

With the introduction of the correlated latent code $z_R$, fair fine-grained image editing—as a fundamental concept in counterfactual fairness—can be naturally achieved by aligning latent codes from different samples. We use linear interpolation to synthesize a latent code:
$z' = (1-\lambda) z_1 + \lambda z_2$,
where $z_1$ is the latent code from the source image and $z_2$ is the corresponding code from the reference image. The synthesized $z'$ replaces $z_1$, enabling a gradual transfer from the source to the reference latent code.

In our experiments, following the setup in the previous section where \textit{Smiling} is the target label $Y$ and \textit{Gender} is the sensitive attribute $S$, we generate interpolated latent codes between source and reference images. In the blue-framed subfigure of Figure~\ref{fig:sup_interpolation}, images are generated by interpolating $z_Y$ and $z_S$: the horizontal axis shows the transition of $z_S$ from the source to the reference image, while the vertical axis shows the corresponding change in $z_Y$. During this interpolation, $z_R$ and $z_X$ remain unchanged, with the interpolation parameters for both $z_Y$ and $z_S$ set to $\lambda \in \{0, 0.33, 0.66, 1\}$.

Similarly, in the red-framed subfigure, images are generated by interpolating $z_Y$ and $z_R$. Here, the horizontal axis corresponds to the transition of $z_R$ from the source to the reference image, and the vertical axis corresponds to $z_Y$. Note that the source image’s $z_S$ is fully replaced by that of the reference image, and $z_X$ remains constant. Initially, the interpolation parameters for $z_R$ are set to $\lambda \in \{0.5, 1\}$, and when combined with the final column of the blue-framed subfigure, the range is extended to $\lambda \in \{0, 0.5, 1\}$.

Figure~\ref{fig:sup_interpolation} demonstrates a smooth transformation of each attribute, with modifications in one latent code minimally affecting the others, a key characteristic of effective disentanglement. Specifically, as the correlated latent code $z_R$ captures sensitive relevant information, we can explicitly control these properties: in the left subfigure, we gradually introduce makeup (such as enhanced lipstick and eyeshadow), while in the right subfigure, we progressively add a mustache. More fair fine-grained image editing experiment results can be seen in \textit{Appendix 7}. 

Similarly, we measure $\Delta \text{FID}$ and $\Delta \text{IS}$ to quantitatively assess the quality of fine-grained image editing. Unlike the evaluation setup in the previous section, we compute the differences in evaluation metrics between the direct reconstruction of the input image and the reconstructions obtained through latent code traversals for each $\lambda$ combination. Detailed experimental settings are provided in \textit{Appendix 6}.
Table~\ref{tb:quantitative_edit} summarizes the comparison results, showing that our method exhibits both lower $\Delta \text{FID}$ and $\Delta \text{IS}$ values compared to other fair generation methods. These results indicate that our fine-grained image editing approach not only ensures smoother attribute transformations and superior image fidelity, but also allows for more precise control of task-relevant features.

\subsection{Fair Text-to-Image Editing}

To further validate the capability and explore the applicability of our method, we integrated it as an adaptor on top of a pre-trained, frozen CLIP image encoder \cite{radford2021learningtransferablevisualmodels,xiao2025prompt,xiao2025via,xiao2025visual} and trained it on Facet dataset~\cite{gustafson2023facetfairnesscomputervision} to enhance fairness in vision-language tasks. Table~\ref{tb:main_CLIP_facet} presents the experimental results. These results demonstrate that our approach significantly improves fairness without compromising performance compared to the linear probing baseline (ERM), underscoring its potential for a range of vision-language tasks with fairness considerations.

Furthermore, we applied our method in StyleCLIP~\cite{Patashnik_2021_ICCV} as a fair discriminator to address inherent fairness issues, such as career-gender biases, which persist even when an identity preservation loss is employed. As illustrated in Figure~\ref{fig:StyleCLIP}, StyleCLIP~\cite{Patashnik_2021_ICCV} exhibits a bias by correlating the role of ``dancer" with a specific gender. In contrast, our method effectively mitigates this bias while maintaining the efficacy of attribute modification. See \textit{Appendix 7} for details.

\begin{table}[h]
\centering
\caption{\textbf{FID and IS difference between original reconstruction and traversed target/sensitive codes' reconstruction.}}
\label{tb:quantitative_edit}
\vspace{-1em}
\renewcommand{\arraystretch}{1} 
\setlength{\tabcolsep}{3pt} 
\begin{adjustbox}{width=\columnwidth,center}
\begin{tabular}{c||cccccc} 
\Xhline{2\arrayrulewidth}
\rowcolor{gray!20}
& \textbf{CAD-VAE} & FADES & GVAE & FFVAE & ODVAE & FairFactorVAE \\ 
\hline
$\Delta \textit{\text{FID}} \downarrow$ & \textbf{1.642} & \textcolor{red}{2.362} & 4.023 & 2.789 & 15.893 & 7.120  \\ 
\hline
$\Delta \textit{\text{IS}} \downarrow$ & \textbf{1.849} & \textcolor{red}{2.919} & 4.848 & 5.292 & 6.890 & 5.767 \\ 
\Xhline{2\arrayrulewidth}
\end{tabular}
\end{adjustbox}
\vspace{-1.0em}
\end{table}

\begin{figure}[h]

    \centering
    \begin{subfigure}[b]{0.32\linewidth}
        \centering
        \includegraphics[width=\linewidth]{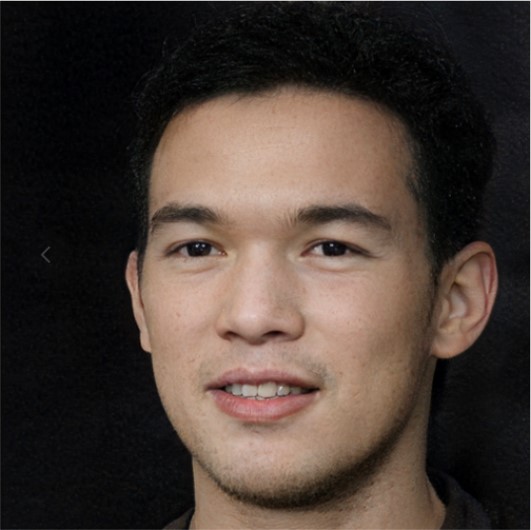}
        \caption{Original}
        \label{fig:sub1}
    \end{subfigure}%
    \hfill
    \begin{subfigure}[b]{0.32\linewidth}
        \centering
        \includegraphics[width=\linewidth]{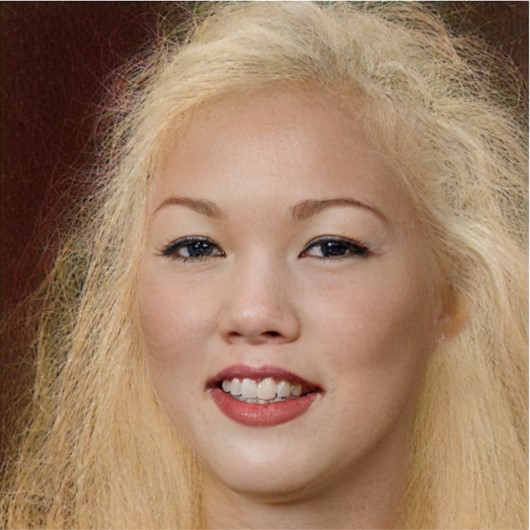}
        \caption{StyleCLIP}
        \label{fig:sub2}
    \end{subfigure}%
    \hfill
    \begin{subfigure}[b]{0.32\linewidth}
        \centering
        \includegraphics[width=\linewidth]{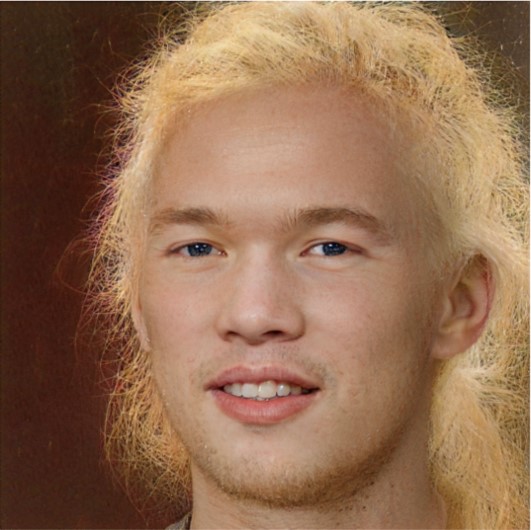}
        \caption{StyleCLIP+}
        \label{fig:sub3}
    \end{subfigure}
    \vspace{-1em}
    \caption{\textbf{Style transfer using StyleCLIP and the CAD-VAE extension.} This example transforms the (a) into ``a dancer with long blonde hair." ``StyleCLIP+" means StyleCLIP + CAD-VAE.}
    \label{fig:StyleCLIP}
    \vspace{-2em}

\end{figure}

\begin{table}[h]
\centering
\scriptsize 
\caption{\textbf{Performance of CLIP(ViTB/32) on Facet dataset.} WG: Worst Group, Gap: Difference between WG and Avg.}
\vspace{-1.0em}
\label{tb:main_CLIP_facet}
\renewcommand{\arraystretch}{0.8} 
\setlength{\tabcolsep}{2pt}      
\begin{adjustbox}{width=\columnwidth,center}
\begin{tabular}{c||ccc|ccc} 
\Xhline{2\arrayrulewidth}
\rowcolor{gray!20}
\textbf{Method} & \multicolumn{3}{c|}{\textbf{Top-1 Acc. (\%)}} & \multicolumn{3}{c}{\textbf{Top-3 Acc. (\%)}} \\ 
\rowcolor{gray!10}
& WG $\uparrow$ & Avg $\uparrow$ & Gap $\downarrow$ & WG $\uparrow$ & Avg $\uparrow$ & Gap $\downarrow$ \\ 
\Xhline{2\arrayrulewidth}
Zero-shot   & 2.79 & 53.45 & 50.66 & 15.31 & 76.79 & 61.48 \\ 
Linear prob & 1.17 & 65.46 & 64.29 & 1.79  & 85.34 & 83.55 \\ 
CAD-VAE       & \textbf{69.97} & \textbf{70.54} & \textbf{0.57} & \textbf{85.36} & \textbf{85.95} & \textbf{0.59} \\ 
\Xhline{2\arrayrulewidth}
\end{tabular}
\end{adjustbox}
\vspace{-1.0em}
\end{table}


\section{Conclusion}
\label{sec:conclusion}

Our method aims to solve fairness concerns in representation learning and deep generative models. By introducing a correlated latent code that captures shared information, sensitive information leakage can be eliminated directly and efficiently without conflicting with the prediction objective, which is a core issue in disentanglement, by minimizing the conditional mutual information between target latent code and sensitive latent code. Parallel with our explicit relevance learning strategy imposed on the correlated latent code, it is encouraged to capture the essential shared information that cannot be perfectly separated without additional domain knowledge. Various benchmark tasks further demonstrate the robustness and wide applicability of our method. 

\clearpage

\section*{Acknowledgment}

This manuscript was co-authored by Oak Ridge National Laboratory (ORNL), operated by UT-Battelle, LLC under Contract No. DE-AC05-00OR22725 with the U.S. Department of Energy. Any subjective views or opinions expressed in this paper do not necessarily represent those of the U.S. Department of Energy or the United States Government.



\bibliography{aaai2026}
\clearpage
\section*{Appendix}

\newtheorem{remark}{Remark}

\section{Relation between Conditional Mutual Information and Conditional Independence}
\label{sec:appendix_section_A}

\begin{proposition}
\label{prop:CMI_Independence}
Let \(X, Y, Z\) be random variables with a joint distribution \(P_{X,Y,Z}\). Then the following are equivalent:
\begin{enumerate}
    \item \(I(X; Y \mid Z) = 0\),
    \item \(X\) and \(Y\) are conditionally independent given \(Z\), i.e., \(X \perp Y \mid Z\).
\end{enumerate}
\end{proposition}

\textbf{(2) \(\implies\) (1).} 
If \(X\) is conditionally independent of \(Y\) given \(Z\), we have
\begin{equation}
    P_{X,Y|Z}(x,y \mid z) \;=\; P_{X|Z}(x\mid z)\,P_{Y|Z}(y\mid z).
\end{equation}
Thus, for each \(z\) with \(P_Z(z) > 0\),
\begin{equation}
    \log\!\biggl[\frac{P_{X,Y|Z}(x,y\mid z)}{P_{X|Z}(x\mid z)\,P_{Y|Z}(y\mid z)}\biggr]
\;=\;\log(1)
\;=\;0.
\end{equation}
Therefore,
\begin{equation}
\begin{aligned}
&I(X;Y \!\mid\! Z) \\ \!
\!&=\! \!\sum_{z} \!P_Z(z)\! 
\sum_{x,y}\! P_{X,Y \mid Z}(x,y \!\mid\! z) \!
\log\!\biggl[\frac{P_{X,Y|Z}(x,y \!\mid\! z)}{P_{X|Z}(x\!\mid\! z)\,P_{Y|Z}(y\!\mid\! z)}\!\biggr]  \\ \!
&= 0.
\end{aligned}
\end{equation}

\noindent
\textbf{(1) \(\implies\) (2).}
Suppose \(I(X;Y \mid Z) = 0\). Define
\begin{equation}
    F(x,y) \;:=\; \frac{P_{X|Z}(x \mid z)\,P_{Y|Z}(y \mid z)}{P_{X,Y|Z}(x,y \mid z)},
\end{equation}
for each \(z\) such that \(P_Z(z) > 0\). A direct calculation shows
\begin{equation}
\begin{aligned}
&\sum_{x,y} P_{X,Y|Z}(x,y\mid z)\;F(x,y) \\
&\!=\!\sum_{x,y} 
     P_{X|Z}(x\mid z)\,
     P_{Y|Z}(y\mid z)
\!=\!1.
\end{aligned}
\end{equation}
The conditional mutual information can be written as
\begin{equation}
\begin{aligned}
&I(X;Y \!\mid\! Z) \\ \!
\!&=\! \!\sum_{z} \!P_Z(z)\! 
\sum_{x,y}\! P_{X,Y \mid Z}(x,y \!\mid\! z) \!
\log\!\biggl[\frac{P_{X,Y|Z}(x,y \!\mid\! z)}{P_{X|Z}(x\!\mid\! z)\,P_{Y|Z}(y\!\mid\! z)}\!\biggr]  \\ \!
&= -\,\sum_{z} P_{Z}(z)\sum_{x,y}P_{X,Y|Z}(x,y\mid z)\,\log F(x,y).
\end{aligned}
\end{equation}

Using the identity
\begin{equation}
\!-\!\,\!\log F(x,y) 
\!=\!\bigl[F(x,y) \! - \! 1 \! - \! \log F(x,y)\bigr]
\!-\!\bigl[F(x,y) \! - \! 1\bigr],
\end{equation}

and rearranging, one obtains
\begin{equation}
\label{I_reduced}
    \begin{aligned}
        &I(X;Y\mid Z) \\
\!&= \!\sum_{z}\! P_Z(z)\!\sum_{x,y} \!P_{X,Y|Z}(x,y\!\mid\! z)\!\Bigl[F(x,y)\! - \!1 \!- \!\log \!F(x,y)\Bigr]\!.
    \end{aligned}
\end{equation}

To see why this rearrangement holds, substitute the above identity into 
\(\sum_{x,y}P_{X,Y|Z}(x,y\mid z)\,\log F(x,y)\). The result is a difference of two sums; one of these sums,
\(\sum_{x,y} P_{X,Y|Z}(x,y\mid z)\,\bigl[F(x,y) - 1\bigr]\),
is zero because
\begin{equation}
    \begin{aligned}
        &\sum_{x,y} P_{X,Y|Z}(x,y\mid z)\,\bigl[F(x,y) - 1\bigr] \\
&\;=\;
\sum_{x,y} P_{X,Y|Z}(x,y\mid z)\,F(x,y)
\!-\!
\sum_{x,y} P_{X,Y|Z}(x,y\mid z) \\
&\;=\; 1 - 1 = 0.
    \end{aligned}
\end{equation}

Hence we arrive at the above expression for \eqref{I_reduced}.

Since \(\log t \le t - 1\) for all \(t>0\), we have
\begin{equation}
    F(x,y) - 1 - \log F(x,y) \;\ge\; 0,
\end{equation}

and hence each summand in the expression for \(I(X;Y\mid Z)\) is nonnegative. Because \(I(X;Y\mid Z)=0\) by hypothesis, it must be that
\begin{equation}
    F(x,y) - 1 - \log F(x,y) \;=\; 0
\quad \text{for all \(x,y\)}, 
\end{equation}

implying \(F(x,y) = 1\). Therefore,
\begin{equation}
    F(x,y) = \frac{P_{X|Z}(x \mid z)\,P_{Y|Z}(y \mid z)}{P_{X,Y|Z}(x,y \mid z)}
\;=\;1,
\end{equation}

so that
\begin{equation}
    P_{X,Y|Z}(x,y \mid z)
\;=\;
P_{X|Z}(x \mid z)\,P_{Y|Z}(y \mid z).
\end{equation}

This is precisely the definition of conditional independence: \(X \perp Y \mid Z\). 

Since we have shown both directions \(\textbf{(2)}\implies\textbf{(1)}\) and \(\textbf{(1)}\implies\textbf{(2)}\), the proof is complete.
\section{Complete Transformation Process of Conditional Mutual Information}
\label{sec:appendix_section_B}

$H_\phi(* \mid *)$ stand the conditional entropy, and
\begin{equation}
\label{sup:predY}
    \hat{Y} = f_y(z_Y, z_R)
\end{equation}
\begin{equation}
\label{sup:predS}
    \hat{S} = f_s(z_S, z_R).
\end{equation}

Incorporating \eqref{sup:predS},
we have:
\begin{equation}
\label{sup:CMI_ys_zr_simplify}
\begin{aligned}
    I_\phi(\hat{Y}; \hat{S} \mid z_R) &= H_\phi(\hat{Y} \mid z_R) - H_\phi(\hat{Y} \mid \hat{S}, z_R) \\
    &= H_\phi(\hat{Y} \mid z_R) - H_\phi(\hat{Y} \mid f_s(z_S, z_R), z_R) \\
    &= H_\phi(\hat{Y} \mid z_R) - H_\phi(\hat{Y} \mid z_S, z_R) \\
    &= I_\phi(\hat{Y}; z_S \mid z_R).
\end{aligned}
\end{equation}

Incorporating \eqref{sup:predY},
we have:
\begin{equation}
\label{sup:CMI_sy_zr_simplify}
\begin{aligned}
    I_\phi(\hat{S}; \hat{Y} \mid z_R) &= H_\phi(\hat{S} \mid z_R) - H_\phi(\hat{S} \mid \hat{Y}, z_R) \\
    &= H_\phi(\hat{S} \mid z_R) - H_\phi(\hat{S} \mid f_y(z_Y, z_R), z_R) \\
    &= H_\phi(\hat{S} \mid z_R) - H_\phi(\hat{S} \mid z_Y, z_R) \\
    &= I_\phi(\hat{S}; z_Y \mid z_R).
\end{aligned}
\end{equation}

Therefore, with the introduction of the correlated latent code $z_R$ that captures all relevant information between $\hat{Y}$ and $\hat{S}$:
\begin{equation}
\label{sup:CMI_equal}
\begin{aligned}
\min_{\phi} 
\Bigl[
    &I_\phi(\hat{Y}; \hat{S} \mid z_R) + I_\phi(\hat{S}; \hat{Y} \mid z_R)
\bigr] \\
    &\equiv 
\min_{\phi} 
\Bigl[
    I_\phi(\hat{Y}; z_S \mid z_R) + I_\phi(\hat{S}; z_Y \mid z_R)
\Bigr].
\end{aligned}
\end{equation}

For the minimization of $I_\phi(\hat{Y}; z_S \mid z_R)$, as shown in \eqref{sup:CMI_ys_zr_simplify}, since $z_R$ is given as a condition, we can consider this CMI formula as a function where the independent variable is $z_S$ and the dependent variable is $\hat{Y}$, as shown as:
\begin{equation}
\label{sup:CMI_func_y_zs}
\mathcal{L}_{\hat{Y}}(z_S) = H_\phi(\hat{Y} \mid z_R) - H_\phi(\hat{Y} \mid z_S, z_R), 
\end{equation}
where $z_R$ is determined here, $H_\phi(\hat{Y} \mid z_R)$ is a constant, henceforth we need to minimize $-H_\phi(\hat{Y} \mid z_S, z_R)$. Empirically, we directly minimize the lower bound of it: $-H_\phi(\hat{Y} \mid z_S)$, since: $-H_\phi(\hat{Y} \mid z_S, z_R) \geq -H_\phi(\hat{Y} \mid z_S)$.

For the minimization of $I_\phi(\hat{S}; z_Y \mid z_R)$, as shown in \eqref{sup:CMI_sy_zr_simplify}, since $z_R$ is given as a condition, we can consider this CMI formula as a function where the independent variable is $z_Y$ and the dependent variable is $\hat{S}$, as shown as:
\begin{equation}
\label{sup:CMI_func_s_zy}
\mathcal{L}_{\hat{S}}(z_Y) = H_\phi(\hat{S} \mid z_R) - H_\phi(\hat{S} \mid z_Y, z_R), 
\end{equation}
where $z_R$ is determined here, $H_\phi(\hat{S} \mid z_R)$ is a constant, henceforth we need to minimize $-H_\phi(\hat{S} \mid z_Y, z_R)$. Empirically, we directly minimize the lower bound of it: $-H_\phi(\hat{S} \mid z_Y)$, since: $-H_\phi(\hat{S} \mid z_Y, z_R) \geq -H_\phi(\hat{S} \mid z_Y)$.

After these simplifications, we can introduce the CMI loss to minimize \eqref{sup:CMI_equal}:
\begin{equation}
\label{sup:CMI_loss}
\min_{\phi} 
\Bigl[
\mathcal{L}_{\text{CMI}}(\omega_{y\_op}; \omega_{s\_op}; \phi) \Bigr] \!
 \!=\! - (H_\phi(\hat{Y} \mid z_S) + H_\phi(\hat{S} \mid z_Y)),
\end{equation}
where only update encoder parameters $\phi$.

\section{Complete Calculation Formula of Entropy Term mentioned in Main Paper}
\label{sec:appendix_section_C}

1. $H_{\phi}(\hat{Y}\mid z_S)$:
We use the opponent classifier $f_{y\_op}(z_S)$, which predicts $y$ from $z_S$. Denoting
    \[
    p_{\theta}\bigl(\hat{y} \mid z_S^{(k)}\bigr)
    \;=\;
    f_{y\_op}\bigl(z_S^{(k)}\bigr),
    \]
    then
    \begin{equation}
    \begin{aligned}
    &H_{\phi}(\hat{Y}\mid z_S) \\
    &\;=\;
    \mathbb{E}_{q_{\phi}(z_S \mid x)}
    \biggl[
      -\sum_{\,\hat{y} \in \mathcal{Y}}\,p_{\theta}(\hat{y} \mid z_S)\,\log \,p_{\theta}(\hat{y} \mid z_S)
    \biggr] \\
    &\;=\;
    \frac{1}{\lvert B\rvert}\,\sum_{i=1}^{\lvert B\rvert}\,\sum_{\,\hat{y} \in \mathcal{Y}}\,
    \biggl[
      -
      p_{\theta}\bigl(\hat{y} \mid z_S^{(i)}\bigr)
      \log p_{\theta}\bigl(\hat{y} \mid z_S^{(i)}\bigr)
    \biggr].
    \end{aligned}
    \end{equation}
    
    Here, $z_S^{(i)}$ denotes the $z_S$ sample from the $i$-th element in a mini-batch of size $|B|$; the distribution $q_{\phi}(z_S \mid x)$ is given by the encoder.

2. $H_{\phi}(\hat{S}\mid z_Y)$:
Similarly, we employ the opponent classifier $f_{s\_op}(z_Y)$ to measure how much $z_Y$ retains information about $s$. We define
    \[
    p_{\theta}\bigl(\hat{s} \mid z_Y^{(k)}\bigr)
    \;=\;
    f_{s\_op}\bigl(z_Y^{(k)}\bigr),
    \]
    and 
    \begin{equation}
    \begin{aligned}
    &H_{\phi}(\hat{S}\mid z_Y) \\
    &\;=\;
    \mathbb{E}_{q_{\phi}(z_Y \mid x)}
    \biggl[
      -\sum_{\,\hat{s} \in \mathcal{S}}\,
      p_{\theta}(\hat{s} \mid z_Y)
      \,\log \,p_{\theta}(\hat{s} \mid z_Y)
    \biggr] \\
    &\;=\;
    \frac{1}{\lvert B\rvert}\,\sum_{i=1}^{\lvert B\rvert}\,\sum_{\,\hat{s} \in \mathcal{S}}\,
    \biggl[
      -
      p_{\theta}\bigl(\hat{s} \mid z_Y^{(i)}\bigr)
      \log p_{\theta}\bigl(\hat{s} \mid z_Y^{(i)}\bigr)
    \biggr].
    \end{aligned}
    \end{equation}

    Here, $z_Y^{(i)}$ denotes the $z_Y$ sample from the $i$-th element in a mini-batch of size $|B|$; the distribution $q_{\phi}(z_Y \mid x)$ is given by the encoder.

3. $H_\phi(\hat{Y} \mid z_R)$: We approximate $p_\theta(\hat{y} \mid z_R)$ by marginalizing over $z_Y$:
   \begin{equation}
   \begin{aligned}
   p_\theta(\hat{y} \mid z_R^{(k)}) &= \mathbb{E}_{p(x)} \left[ \mathbb{E}_{q_\phi(z_Y \mid x)} \left[ p_\theta(\hat{y} \mid z_Y, z_R^{(k)}) \right] \right] \\ &\approx \frac{1}{|B|}\sum_{i=1}^{|B|} p_\theta(\hat{y} \mid z_Y^{(i)}, z_R^{(k)}), 
   \end{aligned}
   \end{equation}
   where $p_\theta(\hat{y} \mid z_Y^{(i)}, z_R^{(k)})$ is obtained by $f_y(z_Y, z_R)$:
   \begin{equation}
    p_\theta(\hat{y} \mid z_Y^{(i)}, z_R^{(k)})
    \;=\;
    f_y\bigl(z_Y^{(i)}, z_R^{(k)}\bigr), 
    \end{equation}
   then
   \begin{equation}
   \begin{aligned}
   &H_\phi(\hat{Y} \mid z_R) \\ &= \mathbb{E}_{q_\phi(z_R \mid x)}\left[-\sum_{\hat{y} \in \mathcal{Y}} p_\theta(\hat{y} \mid z_R)\log p_\theta(\hat{y} \mid z_R)\right] \\
    &= \frac{1}{|B|} \sum_{i=1}^{|B|} \sum_{\hat{y} \in \mathcal{Y}} \left[ -p_\theta(\hat{y} \mid z_R^{(i)}) \log p_\theta(\hat{y} \mid z_R^{(i)}) \right], 
    \end{aligned}
   \end{equation}

4. $H_\phi(\hat{S} \mid z_R)$: We approximate $p_\theta(\hat{s} \mid z_R)$ by marginalizing over $z_S$:
   \begin{equation}
   \begin{aligned}
   p_\theta(\hat{s} \mid z_R^{(k)}) &= \mathbb{E}_{p(x)} \left[ \mathbb{E}_{q_\phi(z_S \mid x)} \left[ p_\theta(\hat{s} \mid z_S, z_R^{(k)}) \right] \right] \\ &\approx \frac{1}{|B|}\sum_{i=1}^{|B|} p_\theta(\hat{s} \mid z_S^{(i)}, z_R^{(k)}), 
   \end{aligned}
   \end{equation}
   where $p_\theta(\hat{s} \mid z_S^{(i)}, z_R^{(k)})$ is obtained by $f_s(z_S, z_R)$:
   \begin{equation}
    p_\theta(\hat{s} \mid z_S^{(i)}, z_R^{(k)})
    \;=\;
    f_s\bigl(z_S^{(i)}, z_R^{(k)}\bigr), 
    \end{equation}
   then
   \begin{equation}
   \begin{aligned}
   &H_\phi(\hat{S} \mid z_R) \\ &= \mathbb{E}_{q_\phi(z_R \mid x)}\left[-\sum_{\hat{s} \in \mathcal{S}} p_\theta(\hat{s} \mid z_R)\log p_\theta(\hat{s} \mid z_R)\right] \\
    &= \frac{1}{|B|} \sum_{i=1}^{|B|} \sum_{\hat{s} \in \mathcal{S}} \left[ -p_\theta(\hat{s} \mid z_R^{(i)}) \log p_\theta(\hat{s} \mid z_R^{(i)}) \right], 
    \end{aligned}
   \end{equation}

5. $H_\phi(\hat{Y} \mid Y, z_R)$:
$p_{\theta}(\hat{y} \mid z_R, y)$ can be computed for $z_R^{(k)}$ sampled from an instance \( x^{(k)} \in B_y \) as:
\begin{equation}
\begin{aligned}
\label{sup:cal_P_y_zr_y}
p_{\theta} (\hat{y} | z_R^{(k)}, y) &= \mathbb{E}_{p(x\mid Y=y)} \left[ \mathbb{E}_{q_{\phi}(z_Y \mid x)} \left[ p_{\theta} (\hat{y} \mid z_Y, z_R^{(k)}) \right] \right] \\
&\approx \frac{1}{|B_y|} \sum_{i=1}^{|B_y|} p_{\theta} (\hat{y} | z_Y^{(i)}, z_R^{(k)}),
\end{aligned}
\end{equation}
where \( B_y \) denotes a subset of the batch with \( Y = y \). Then the conditional entropy can be computed as: 
\begin{equation}
\begin{aligned}
\label{sup:cal_H_Y_Y_zr}
&H_{\phi} (\hat{Y} \mid Y, z_R)= \\ &\mathbb{E}_{(x,y)\sim \mathcal{D}}\!\left[ \!\mathbb{E}_{q_\phi(z_R \mid x)}\!\left[ \!-\!\sum_{\hat{y} \in \mathcal{Y}} \!p_\theta(\hat{y} \!\mid\! z_R, y)\!\log p_\theta(\hat{y}\! \mid\! z_R, \!y)\right] \! \right] \!\\
&= \frac{1}{|B|}\! \sum_{y \in \mathcal{Y}}\! \sum_{i=1}^{|B_y|}\! \sum_{\hat{y} \in \mathcal{Y}} \left[ -p_{\theta} (\hat{y} | z_R^{(i)}, y) \!\log p_{\theta} (\hat{y} | z_R^{(i)}, y) \right].
\end{aligned}
\end{equation}

6. $H_\phi(\hat{S} \mid S, z_R)$:
$p_{\theta}(\hat{s} \mid z_R, s)$ can be computed for $z_R^{(k)}$ sampled from an instance \( x^{(k)} \in B_s \) as:
\begin{equation}
\begin{aligned}
\label{sup:cal_P_s_zr_s}
p_{\theta} (\hat{s} | z_R^{(k)}, s) &= \mathbb{E}_{p(x\mid S=s)} \left[ \mathbb{E}_{q_{\phi}(z_S \mid x)} \left[ p_{\theta} (\hat{s} \mid z_S, z_R^{(k)}) \right] \right] \\
&\approx \frac{1}{|B_s|} \sum_{i=1}^{|B_s|} p_{\theta} (\hat{s} | z_S^{(i)}, z_R^{(k)}),
\end{aligned}
\end{equation}
where \( B_s \) denotes a subset of the batch with \( S = s \). Then the conditional entropy can be computed as: 
\begin{equation}
\begin{aligned}
\label{sup:cal_H_S_S_zr}
&H_{\phi} (\hat{S} \mid S, z_R)= \\ &\mathbb{E}_{(x,s)\sim \mathcal{D}}\!\left[ \!\mathbb{E}_{q_\phi(z_R \mid x)}\!\left[ \!-\!\sum_{\hat{s} \in \mathcal{S}} \!p_\theta(\hat{s} \!\mid\! z_R, s)\!\log p_\theta(\hat{s}\! \mid\! z_R, \!s)\right] \! \right] \!\\
&= \frac{1}{|B|}\! \sum_{s \in \mathcal{S}}\! \sum_{i=1}^{|B_s|}\! \sum_{\hat{s} \in \mathcal{S}} \left[ -p_{\theta} (\hat{s} | z_R^{(i)}, s) \!\log p_{\theta} (\hat{s} | z_R^{(i)}, s) \right].
\end{aligned}
\end{equation}

\section{Experiment Setup Details}
\label{sec:appendix_section_D}

\subsection{Comparison Methods}
Specifically, we include FairFactorVAE \cite{kim2018disentangling} and FairDisCo \cite{liu2022fair}, both of which leverage invariant learning techniques to encode latent representations that remain independent of sensitive attributes. Additionally, FFVAE \cite{creager2019flexibly} employs a mutual information minimization strategy to effectively disentangle latent subspaces. GVAE \cite{ding2020guided} takes a distinct approach by minimizing information leakage to achieve fair representation learning. ODVAE \cite{sarhan2020fairness}, in contrast, introduces a non-adversarial learning framework that enforces orthogonal priors on the latent subspace, promoting fairness without adversarial training. Finally, FADES \cite{Jang_2024_CVPR} proposes minimizing conditional mutual information to achieve fairness, aligning closely with our method. As for traditional correlation-aware learning that discussed in Section 2.2, since it require additional annotated data to build casual graph, we except them in our experiment. 

To ensure a fair comparison, we adopt a grid search strategy to fine-tune the hyperparameters of all methods, optimizing for Equalized Odds (EOD) \cite{hardt2016equality}. Specifically, for our method, hyperparameters are explored within the ranges: $\lambda_{CMI} \!\in\! [0, 10]$ and $\lambda_{LRI} \!\in\! [0, 100]$. We maintain a consistent architectural setup across all methods, utilizing a ResNet-18 backbone~\cite{he2016deep} with 512 latent dimensions for all tasks.

\subsection{Dimension of each latent code}
In our experiment, we set equal dimensional sizes for \( z_Y, z_S, z_R \in \mathbb{R}^d \), with \( d \) set as a power of 2. Through empirical selection via grid search, we determined \( d = 32 \) for the reported results. Notably, the total latent dimension was fixed at \( \mathbb{R}^{512} \) across all methods. Consequently, in our approach, we allocated  
$z_X \in \mathbb{R}^{512 - 3 \times 32} = \mathbb{R}^{416}$.  
Our empirical findings revealed that setting \( d \) beyond 64 led to a decline in reconstruction performance. This degradation likely stems from an excessive allocation of information to each subspace, ultimately compromising the model’s reconstructive capacity.

\subsection{Hypermeter Analysis}
\begin{figure}[tp]
    \centering
    \includegraphics[width=1\linewidth]{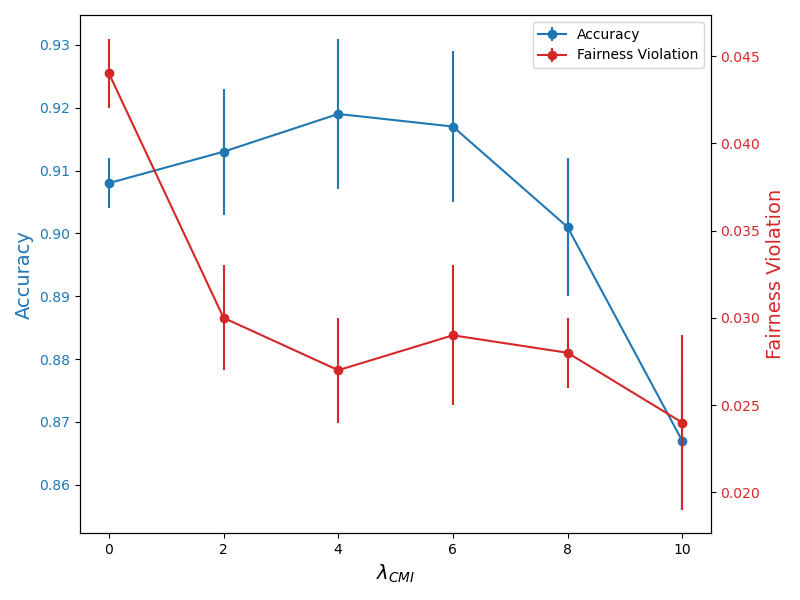}
    \includegraphics[width=1\linewidth]{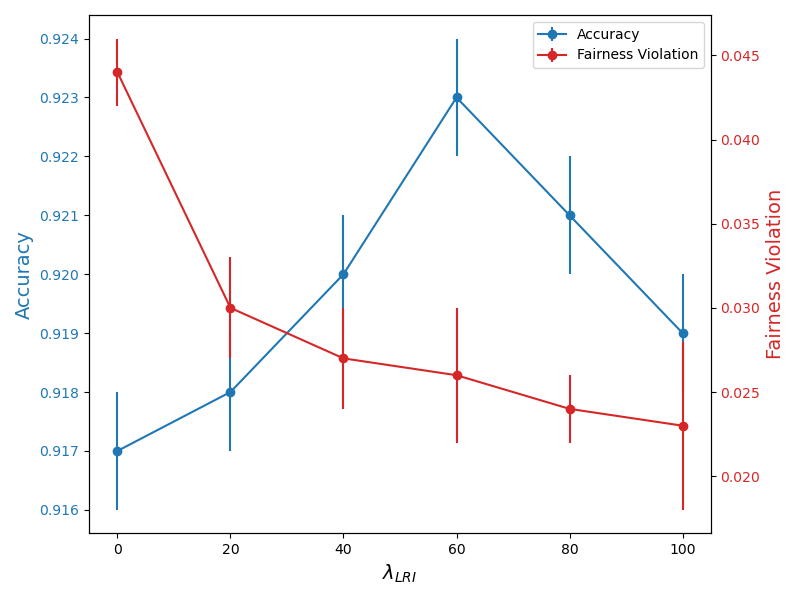}
    \caption{Top: Impact of varying $\lambda_{CMI}$ (with $\lambda_{LRI}$ fixed at 60) on classification accuracy and fairness violation (EOD) in the CelebA dataset classification task. Bottom: Impact of varying $\lambda_{LRI}$ (with $\lambda_{CMI}$ fixed at 5) on the same performance metrics.}
    \label{fig:sup_hyper_ana}
\end{figure}

We evaluated the effect of the two hyperparameters, $\lambda_{CMI} \in [0,10]$ and $\lambda_{LRI} \in [0,100]$, on both classification accuracy and fairness violation (measured as Equalized Odds Difference, EOD) in the CelebA dataset classification task. For the analysis of $\lambda_{CMI}$, we fixed $\lambda_{LRI}$ at 60, while for the analysis of $\lambda_{LRI}$, we fixed $\lambda_{CMI}$ at 5. The result can be seen in Figure~\ref{fig:sup_hyper_ana}.

Specifically, $\lambda_{CMI}$ governs the degree to which the conditional mutual information (CMI) loss is minimized. When $\lambda_{CMI} \leq 5$, we observed a modest increase in accuracy as the hyperparameter increases, accompanied by a uniformly decreasing trend in fairness violation. However, when $\lambda_{CMI} > 5$, further increments in $\lambda_{CMI}$ lead to a decline in accuracy, even though the fairness violation continues to decrease. This behavior suggests that a moderate level of CMI minimization is beneficial in eliminating unwanted information from the latent code, yet an excessive emphasis on minimizing mutual information may inadvertently remove essential target information from $z_Y$.

Similarly, $\lambda_{LRI}$, which regulates the extent to which the latent code $z_R$ captures correlated information, exhibits a comparable pattern. For $\lambda_{LRI} \leq 60$, an increase in $\lambda_{LRI}$ results in an improvement in accuracy and a consistent reduction in fairness violation, indicating that a controlled increase enables $z_R$ to capture sufficient shared patterns while mitigating conflicts between disentanglement and target prediction. Conversely, when $\lambda_{LRI} > 60$, the accuracy begins to decline, despite further reductions in fairness violation. This outcome implies that while a higher $\lambda_{LRI}$ encourages $z_R$ to absorb more correlated information and improves the overall disentanglement performance, it may also lead to the retention of excessive extraneous information, thereby compromising the integrity of the target information.

\subsection{Fair Classification}
The objective of fair classification is to achieve a balance between minimizing fairness violations and maintaining high predictive performance. To evaluate the effectiveness of our proposed method, we conduct experiments on a diverse set of benchmark fairness datasets. For facial attribute classification tasks, we utilize the CelebA \cite{liu2015deep} and UTKFace \cite{zhang2017age} datasets. Following prior works \cite{wang2022fairness, xu2020investigating, zeng2022boosting, Jang_2024_CVPR}, we set the CelebA classification task to predict the "Smiling" attribute, while for UTKFace, the objective is to classify whether a person depicted in the image is over 35 years old, with gender serving as the sensitive attribute. Additionally, the Dogs and Cats dataset \cite{parkhi2012cats} is used to distinguish between dogs and cats, with fur color as the sensitive attribute. Furthermore, we assess fair classification performance using the Colored MNIST dataset \cite{kim2021biaswap, Kim_2019_CVPR, nam2020learning}, which incorporates a controlled color bias in the standard MNIST dataset to simulate spurious correlations. In this dataset, a fixed color is assigned to each digit for a majority of the training samples—specifically, 70\% of the samples have each of the 10 digits correlated with one of 10 predefined colors (with a small random perturbation), while the remaining 30\% are assigned colors uniformly among the other options. For the test set, every digit is paired with a uniformly distributed color assignment to eliminate bias. In our fair classification task, the digit serves as the target attribute and the color as the sensitive attribute. Originally introduced to measure the color bias of classifiers predicting digits, this setup is critical for evaluating a model’s ability to disentangle the target digit information from the spurious color cues.

For model evaluation, we employ a 3-layered Multi-Layer Perceptron (MLP) classifier for all compared methods in all datasets. The input to the classifier consists of target-related features extracted from the pre-trained disentangled representations produced by each method. Specifically, we use $z_Y$ of our method to feed into the classifier, while using $z_X$ from FFVAE \cite{creager2019flexibly} and $z_Y$ from FADES \cite{Jang_2024_CVPR}. For invariant learning methods including FairFactorVAE \cite{kim2018disentangling}, FairDisCo \cite{liu2022fair}, GVAE \cite{ding2020guided}, the entire latent space is utilized for downstream classification tasks. For non-adversarial learning method, we use $z_T$ from ODVAE \cite{sarhan2020fairness}. This experiment setup represents different disentanglement learning strategies as detailed in the related works section.
To assess fairness violations, we use standard metrics including Demographic Parity (DP) \cite{barocas2016big} and Equalized Odds (EOD) \cite{hardt2016equality}. Each method is evaluated over five experimental runs with different random dataset splits when the split is not predefined. This ensures robustness and statistical reliability in performance comparisons.

Demographic Parity (DP) is defined as:
\begin{equation}
    DP = \left| \mathbb{P}(\hat{Y} = 1 \mid A = 0) - \mathbb{P}(\hat{Y} = 1 \mid A = 1) \right|,
\end{equation}
where \(\hat{Y}\) denotes the predicted outcome and \(A\) represents the binary sensitive attribute (e.g., gender, race). This metric captures the absolute difference in positive prediction rates across the groups.

Equalized Odds (EOD) requires that the predictor's true positive rates and false positive rates be equal across groups. It is defined as:
\begin{equation}
\begin{aligned}
    &EOD \!=\! \\ &\max_{y \in \{0,1\}}\! \left| \mathbb{P}(\hat{Y} \!= \!1\! \mid\! Y\! =\! y, A \!= \!0) \!- \!\mathbb{P}(\hat{Y} \!=\! 1 \!\mid\! Y\! =\! y, A\! = \!1) \right|,
\end{aligned}
\end{equation}
where \(Y\) represents the true outcome. This metric considers the maximum discrepancy over both outcome classes, ensuring fairness in both detection and error rates.

\subsection{Fair Counterfactual Generation}
To quantitatively assess the quality of the generated counterfactuals, we compare evaluation metrics between the direct reconstruction of the input image and the reconstructions obtained by randomly permuting the latent representations \(z_Y\) and \(z_S\) within the evaluation set. Specifically, we use the Fréchet Inception Distance (FID)~\cite{heusel2017gans, Jang_2024_CVPR} to assess reconstruction fidelity and the Inception Score (IS)~\cite{chong2020effectively} to evaluate semantic and perceptual quality. Lower values of \(\Delta \text{FID}\) indicate minimal distortion and higher translation quality, while lower values of \(\Delta \text{IS}\) suggest that semantic and perceptual attributes are well preserved. 

We first evaluate reconstruction fidelity by computing the FID between the original image set \(X\) and its reconstructed counterpart \(\hat{X}_{perm}\), where the reconstruction is performed after a random permutation of the latent codes. Formally, we define
\begin{equation}
    \Delta \text{FID} = \text{FID}(X, \hat{X}_{perm}),
\end{equation}
where \(X = \{x_i\}_{i=1}^{N}\) denotes the set of original input images and \(\hat{X}_{perm} = \{\hat{x}_i^{perm}\}_{i=1}^{N}\) represents the corresponding reconstructed images. A lower \(\Delta \text{FID}\) value implies higher translation quality, as it reflects minimal image distortion and fidelity loss during the counterfactual generation process.

In addition, we assess the semantic and perceptual quality by employing the Inception Score. The IS for the direct reconstruction is computed as
\begin{equation}
    \text{IS}(X) = \exp\left(\mathbb{E}_{x\sim p(x)}\left[\text{KL}\left(p(y|x) \,\|\, p(y)\right)\right]\right),
\end{equation}
where \(p(y|x)\) denotes the conditional probability distribution over labels given an image \(x\), and \(p(y)\) represents the marginal distribution over labels. Similarly, we compute \(\text{IS}(\hat{X}_{perm})\) for the permuted reconstructions. The absolute difference between these scores is then defined as
\begin{equation}
    \Delta \text{IS} = \left| \text{IS}(X) - \text{IS}(\hat{X}_{perm}) \right|.
\end{equation}
A lower \(\Delta \text{IS}\) value indicates that the semantic and perceptual attributes of the image are well-preserved after translation, thereby validating the quality of the counterfactuals generated.

Together, these complementary metrics provide a comprehensive evaluation of the counterfactual generation quality. By quantifying both fidelity and semantic consistency, our experimental design rigorously validates the performance of the proposed method.

\subsection{Fair Fine-Grained Image Editing}
Similarly, we quantitatively assess the quality of fine-grained image editing by computing the differences in evaluation metrics between the direct reconstruction of the input image and the reconstructions obtained via latent code traversals with varying \(\lambda\) combinations. Specifically, we define
\begin{equation}
    \Delta \text{FID} = \text{FID}(X, \hat{X}_{trav}),
\end{equation}
and
\begin{equation}
    \Delta \text{IS} = \left| \text{IS}(X) - \text{IS}(\hat{X}_{trav}) \right|,
\end{equation}
where \(X = \{x_i\}_{i=1}^{N}\) denotes the set of original input images, and \(\hat{X}_{trav} = \{\hat{x}_i^{trav}\}_{i=1}^{N}\) represents the corresponding reconstructed images obtained via latent code traversals.

In our experimental setup, where \textit{Smiling} is the target attribute \(Y\) and \textit{Gender} is the sensitive attribute \(S\), we generate interpolated latent codes between source and reference images. Specifically, the interpolation is performed on the latent codes \(z_Y\), \(z_S\), and \(z_R\). The interpolation parameters for \(z_Y\) and \(z_S\) are set to 
\[
\lambda \in \{0,\, 0.33,\, 0.66,\, 1\},
\]
while for \(z_R\) the parameters are chosen from 
\[
\lambda \in \{0,\, 0.5,\, 1\}.
\]
Thus, the overall combination of latent modifications involves \(4 \times 4 \times 3\) distinct \(\lambda\) combinations, corresponding to different configurations for \(z_Y\), \(z_S\), and \(z_R\). During each traversal, the latent code \(z_X\) remains unchanged, ensuring that the variations reflect only the task-relevant attribute transformations. 

In this context, \(\Delta \text{FID}\) measures the reconstruction fidelity, with lower values indicating that the traversed reconstructions closely resemble the original images in terms of their distribution. Likewise, \(\Delta \text{IS}\) quantifies the preservation of semantic and perceptual quality; lower values suggest that the intrinsic content and visual features remain consistent despite the latent manipulations.

\section{Additional Experiment}
\label{sec:appendix_section_E}

\subsection{Fair Classification}
\begin{table*}[htbp]
\centering
\caption{\textbf{Evaluation of downstream classification tasks on CelebA dataset for various target attributes from learned representation.} Best in \textbf{bold}, second in \textcolor{red}{red}.}
\label{sup_tab_fair_cls}
\renewcommand{\arraystretch}{0.9} 
\setlength{\tabcolsep}{3pt} 
\begin{adjustbox}{width=\textwidth,center}
\begin{tabular}{l||ccc|ccc|ccc|ccc} 
\Xhline{2\arrayrulewidth}
\rowcolor{gray!20} \multicolumn{13}{c}{\textbf{CelebA Classification Performance}} \\  
\rowcolor{gray!20}  
\textbf{Methods} & \multicolumn{3}{c|}{Smiling} & \multicolumn{3}{c|}{Blond Hair} & \multicolumn{3}{c|}{Attractive} & \multicolumn{3}{c}{Young} \\  
\rowcolor{gray!20}  
& Acc $\uparrow$ & EOD $\downarrow$ & DP $\downarrow$ & Acc $\uparrow$ & EOD $\downarrow$ & DP $\downarrow$ & Acc $\uparrow$ & EOD $\downarrow$ & DP $\downarrow$ & Acc $\uparrow$ & EOD $\downarrow$ & DP $\downarrow$ \\ 
\Xhline{1.5\arrayrulewidth} 

FADES~\cite{Jang_2024_CVPR} \scriptsize{[CVPR'24]} & 0.918 & \textcolor{red}{0.034} & 0.135 & 0.930 & \textcolor{red}{0.118} & \textcolor{red}{0.153} & 0.763 & \textcolor{red}{0.308} & 0.346 & 0.835 & 0.164 & 0.169 \\ 
GVAE~\cite{ding2020guided} \scriptsize{[CVPR'20]} & \textcolor{red}{0.919} & 0.047 & 0.131 & \textbf{0.940} & 0.484 & 0.247 & \textbf{0.779} & 0.564 & 0.431 & \textcolor{red}{0.841} & 0.209 & 0.176 \\ 
FFVAE~\cite{creager2019flexibly} \scriptsize{[PMLR'19]} & 0.892 & 0.076 & \textcolor{red}{0.072} & 0.926 & 0.301 & 0.201 & 0.749 & 0.359 & \textcolor{red}{0.310} & 0.832 & \textcolor{red}{0.281} & \textbf{0.195} \\ 
ODVAE~\cite{sarhan2020fairness} \scriptsize{[ECCV'20]} & 0.886 & 0.039 & 0.103 & 0.896 & 0.465 & 0.210 & 0.719 & 0.551 & 0.438 & 0.827 & 0.297 & 0.262 \\ 
FairDisCo~\cite{liu2022fair} \scriptsize{[KDD'22]} & 0.839 & 0.074 & \textbf{0.051} & 0.916 & 0.465 & 0.234 & 0.750 & 0.515 & 0.411 & 0.839 & 0.226 & 0.196 \\ 
FairFactorVAE~\cite{liu2023fair} & 0.914 & 0.055 & 0.136 & 0.918 & 0.326 & 0.174 & 0.709 & 0.459 & 0.350 & 0.827 & 0.296 & 0.318 \\ 
\rowcolor{gray!15} \textbf{CAD-VAE (Ours)} & \textbf{0.923} & \textbf{0.021} & 0.112 & \textcolor{red}{0.939} & \textbf{0.105} & \textbf{0.137} & \textcolor{red}{0.773} & \textbf{0.268} & \textbf{0.290} & \textbf{0.847} & \textbf{0.151} & \textcolor{red}{0.155} \\ 
\Xhline{2\arrayrulewidth}
\end{tabular}
\end{adjustbox}
\vspace{-1.5em}
\end{table*}
To further validate the efficacy of our method in achieving fairness via effective disentanglement, we conducted additional experiments with different target attributes. Following the experimental setup in the main paper on the CelebA dataset, our evaluations extended beyond the "Smiling" attribute to include other widely adopted target labels, such as "Blond Hair", "Attractiveness" and "Young" under gender bias conditions. As summarized in Table~\ref{sup_tab_fair_cls}, our method consistently achieves competitive accuracy while significantly mitigating fairness violations compared to existing approaches. These results further substantiate the robustness of our fair disentanglement learning strategy and demonstrate its applicability across a variety of target attributes.

\subsection{Fair Counterfactual Generation}
\begin{figure*}[tp]
    \centering
    \includegraphics[width=1\linewidth]{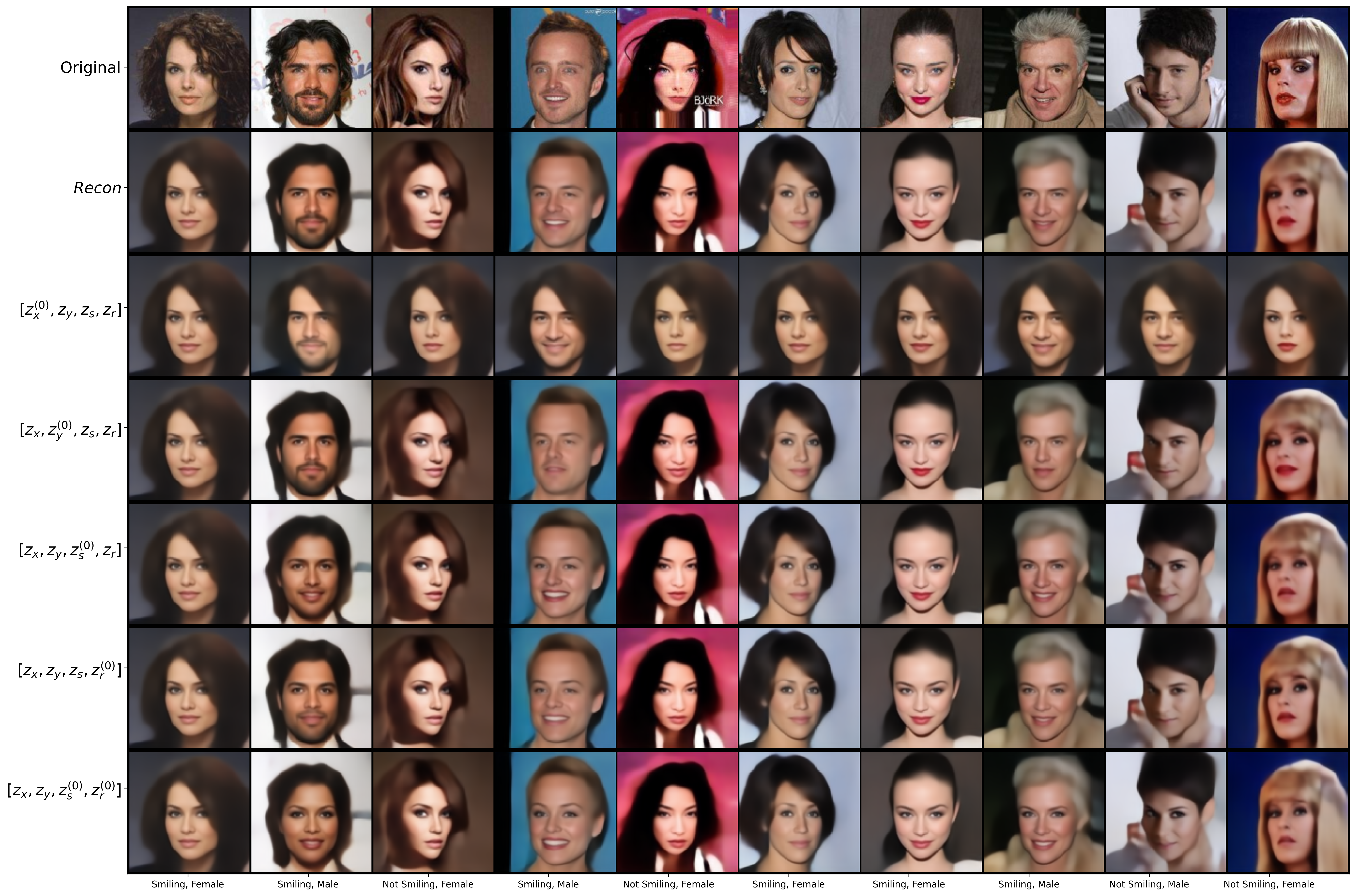}
    \caption{\textbf{Examples of Fair Counterfactual Generation}. The image in the 0th column serves as the reference, while the images in the remaining columns are the source images.}
    \label{fig:sup_counterfactual_1}
\end{figure*}

\begin{figure*}[tp]
    \centering
    \includegraphics[width=1\linewidth]{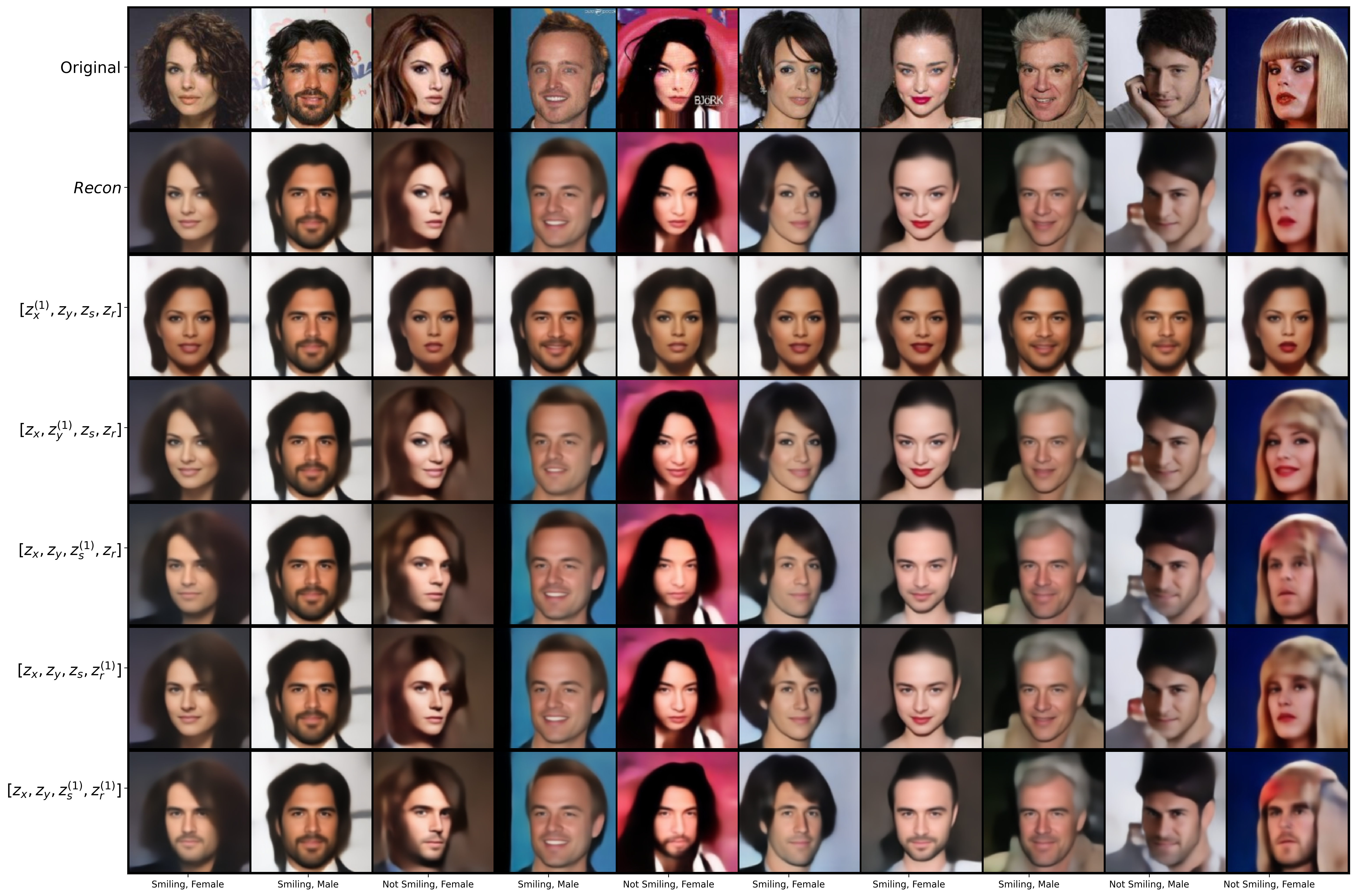}
    \caption{\textbf{Examples of Fair Counterfactual Generation}. The image in the 1th column serves as the reference, while the images in the remaining columns are the source images.}
    \label{fig:sup_counterfactual_2}
\end{figure*}

In this subsection, we present the results of counterfactual generation on the CelebA dataset\cite{liu2015deep}. Figure~\ref{fig:sup_counterfactual_1} and Figure~\ref{fig:sup_counterfactual_2} display a grid of images, each row corresponding to a specific configuration of latent codes. The second row shows the direct reconstruction of the original input image, which is shown in the first row. The intermediate rows illustrate variations derived from different latent code substitutions. For instance, in the third row of Figure~\ref{fig:sup_counterfactual_1}, the latent code configuration \([z_X^{(0)}, z_Y, z_S, z_R]\) is used, where the digit 0 indicates that the reference image is the one in the 0th column. In this configuration, the images in the subsequent columns are generated by replacing their latent code \(z_X\) with the \(z_X\) from the 0th column image, while the remaining latent codes (\(z_Y\), \(z_S\), and \(z_R\)) are retained. This approach effectively alters the corresponding features in the generated images. Our experiments demonstrate that both sensitive and target attributes can be distinctly translated. Moreover, the method supports the simultaneous translation of these attributes, thereby providing greater flexibility in generating counterfactuals. For instance, given an image of a smiling female, our approach can produce images representing a non-smiling male, a non-smiling female, and a smiling male, contributing to improved individual fairness~\cite{jung2025counterfactually} by ensuring that individuals with similar characteristics but different sensitive attributes receive comparable outcomes.

\begin{figure*}[h]
    \centering
    \includegraphics[width=1\linewidth]{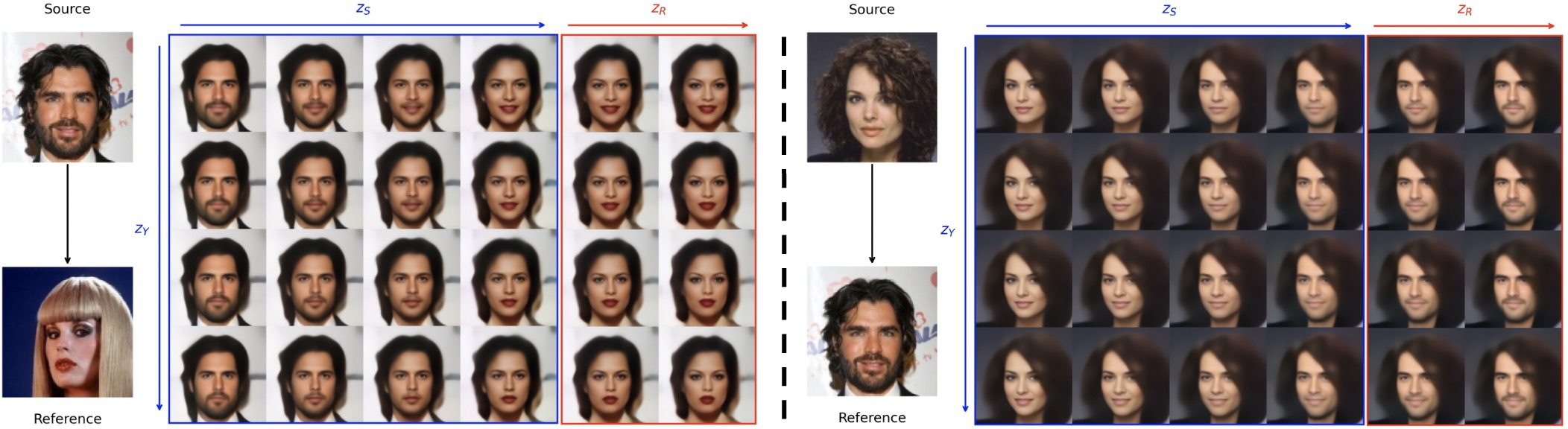}
    \includegraphics[width=1\linewidth]{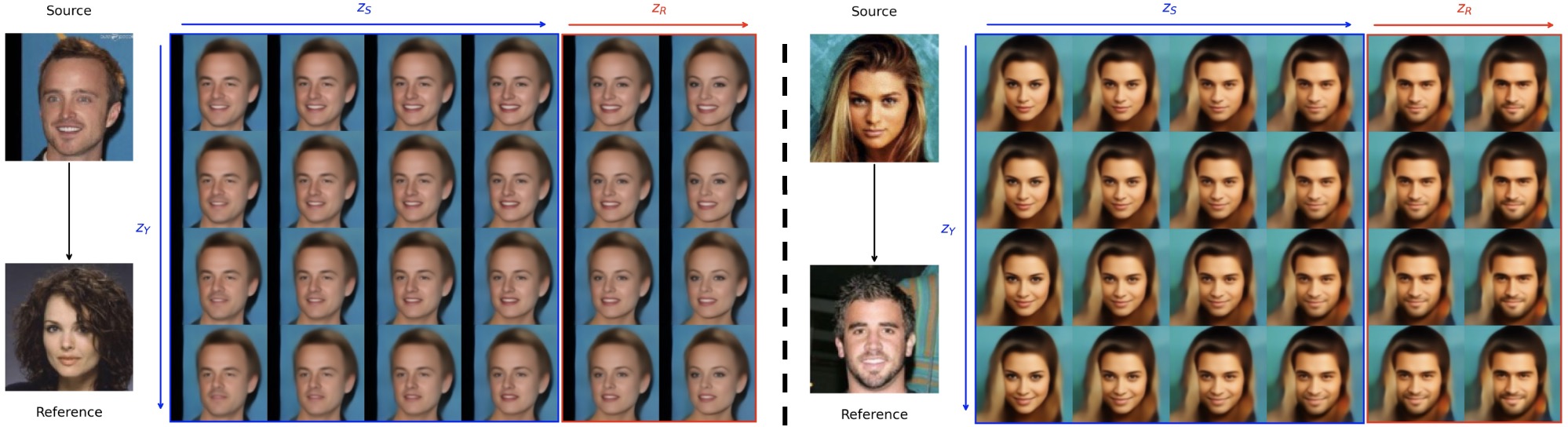}
    \vspace{-1em}
    \caption{\textbf{Examples of Fair Fine-Grained Image Editing}. Zoom in to check. The leftmost column shows the source and reference images. The blue-framed section displays images generated by interpolating \textcolor{blue}{$z_Y$} and \textcolor{blue}{$z_S$} (with \textcolor{red}{$z_R$} and $z_X$ fixed), where the horizontal axis varies \textcolor{blue}{$z_S$} and the vertical axis varies \textcolor{blue}{$z_Y$}. The red-framed section illustrates images produced by interpolating \textcolor{blue}{$z_Y$} and \textcolor{red}{$z_R$} (with \textcolor{blue}{$z_S$} fully replaced by the reference and $z_X$ constant). Modification in one latent code minimally affecting others, harness \textcolor{red}{$z_R$} to edit sensitive relevant feature(makeup or mustache).}
    \label{fig:sup_interpolation}
    \vspace{-1.5em}
\end{figure*}

\subsection{Fair Fine-Grained Image Editing}
In this subsection, we display more fair fine-grained image editing results in Figure~\ref{fig:sup_interpolation}, following the experiment setup in the main paper. Specifically, as the correlated latent code $z_R$ captures sensitive relevant information, we can explicitly control these properties: in the left subfigure, we gradually introduce makeup (such as enhanced lipstick and eyeshadow), while in the right subfigure, we progressively add a mustache.

\begin{figure*}[tp]
    \centering
    \begin{subfigure}[b]{0.49\linewidth}
         \centering
         \includegraphics[width=\linewidth]{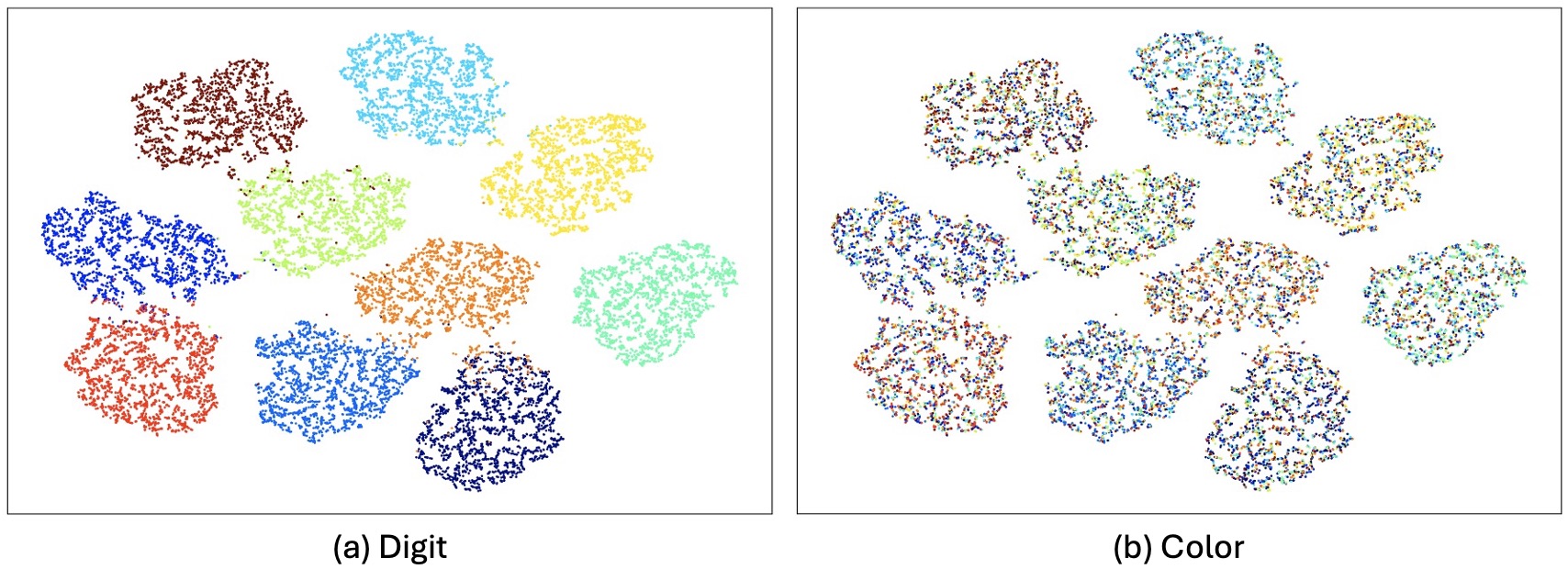}
         \caption{CAD-VAE}
         \label{fig:sub1}
    \end{subfigure}
    \hfill
    \begin{subfigure}[b]{0.49\linewidth}
         \centering
         \includegraphics[width=\linewidth]{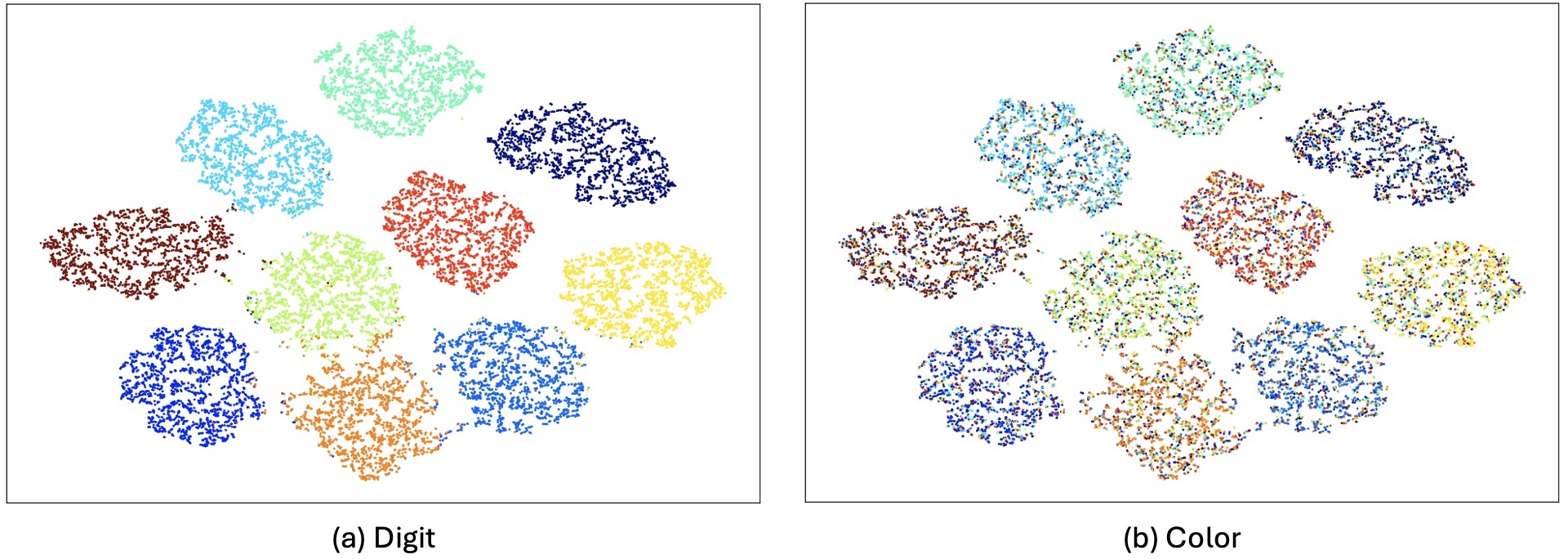}
         \caption{FADES~\cite{Jang_2024_CVPR}}
         \label{fig:sub2}
    \end{subfigure}
    
    \vspace{1em} 

    \begin{subfigure}[b]{0.49\linewidth}
         \centering
         \includegraphics[width=\linewidth]{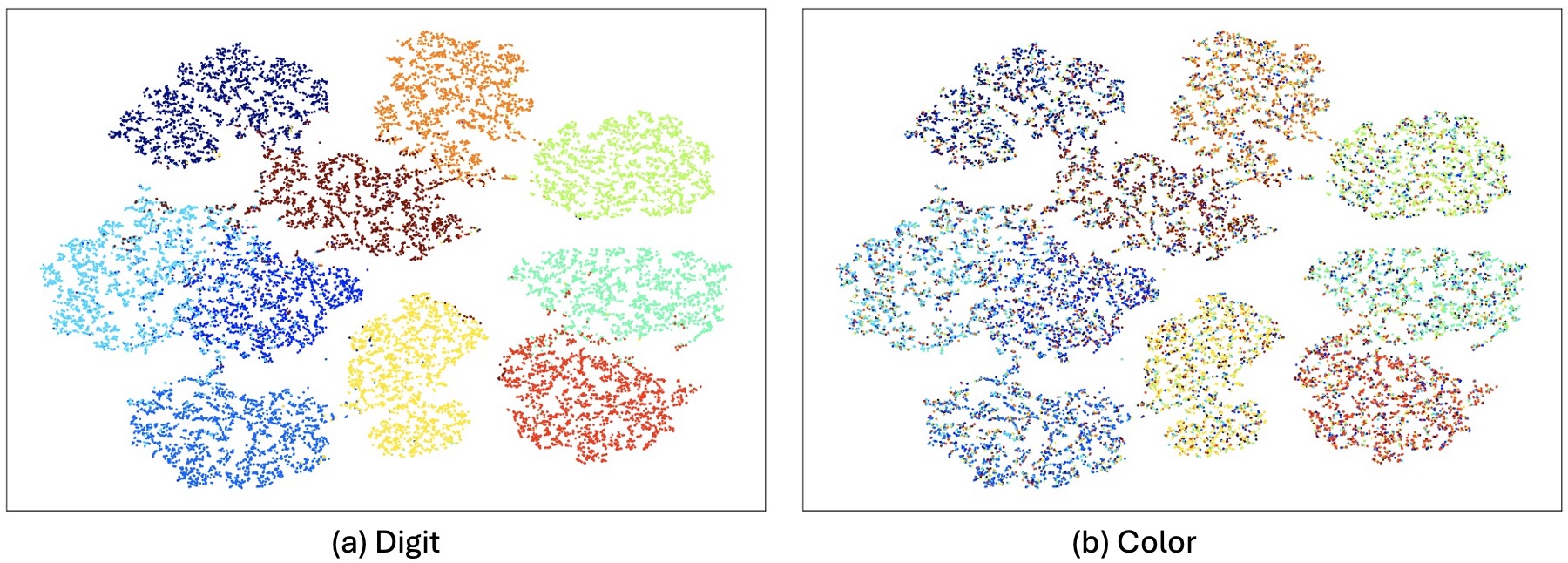}
         \caption{GVAE~\cite{ding2020guided}}
         \label{fig:sub3}
    \end{subfigure}
    \hfill
    \begin{subfigure}[b]{0.49\linewidth}
         \centering
         \includegraphics[width=\linewidth]{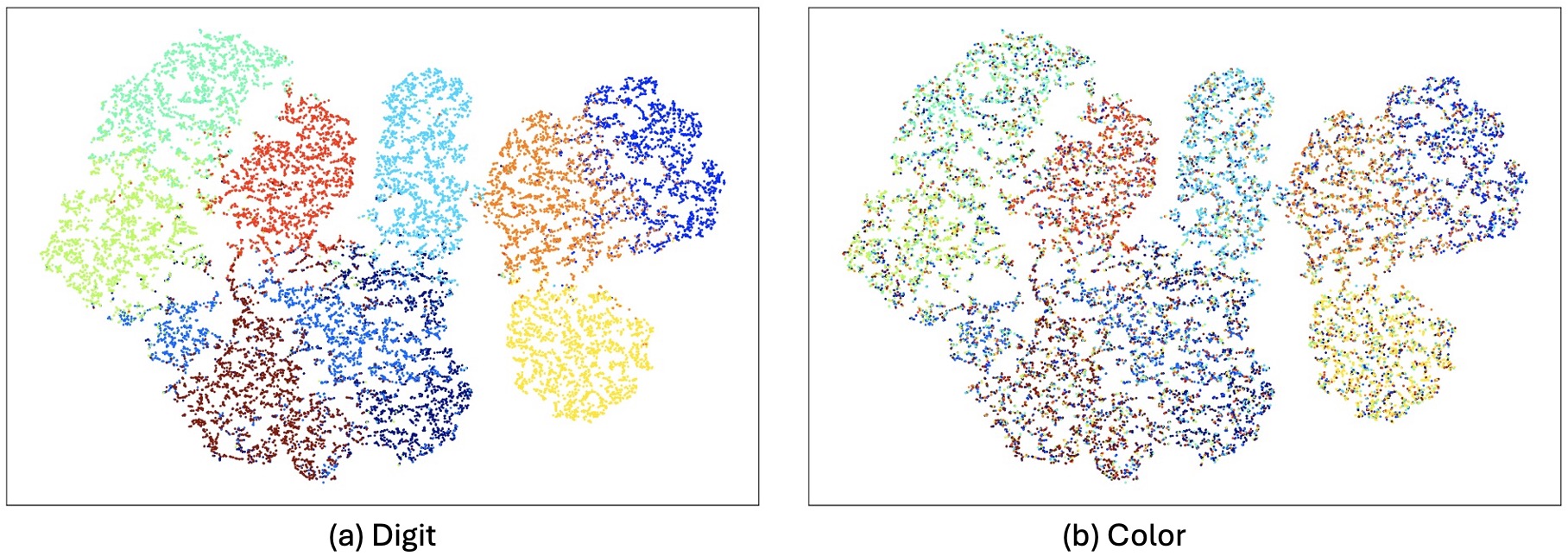}
         \caption{FFVAE~\cite{creager2019flexibly}}
         \label{fig:sub4}
    \end{subfigure}
    
    \vspace{1em}

    \begin{subfigure}[b]{0.49\linewidth}
         \centering
         \includegraphics[width=\linewidth]{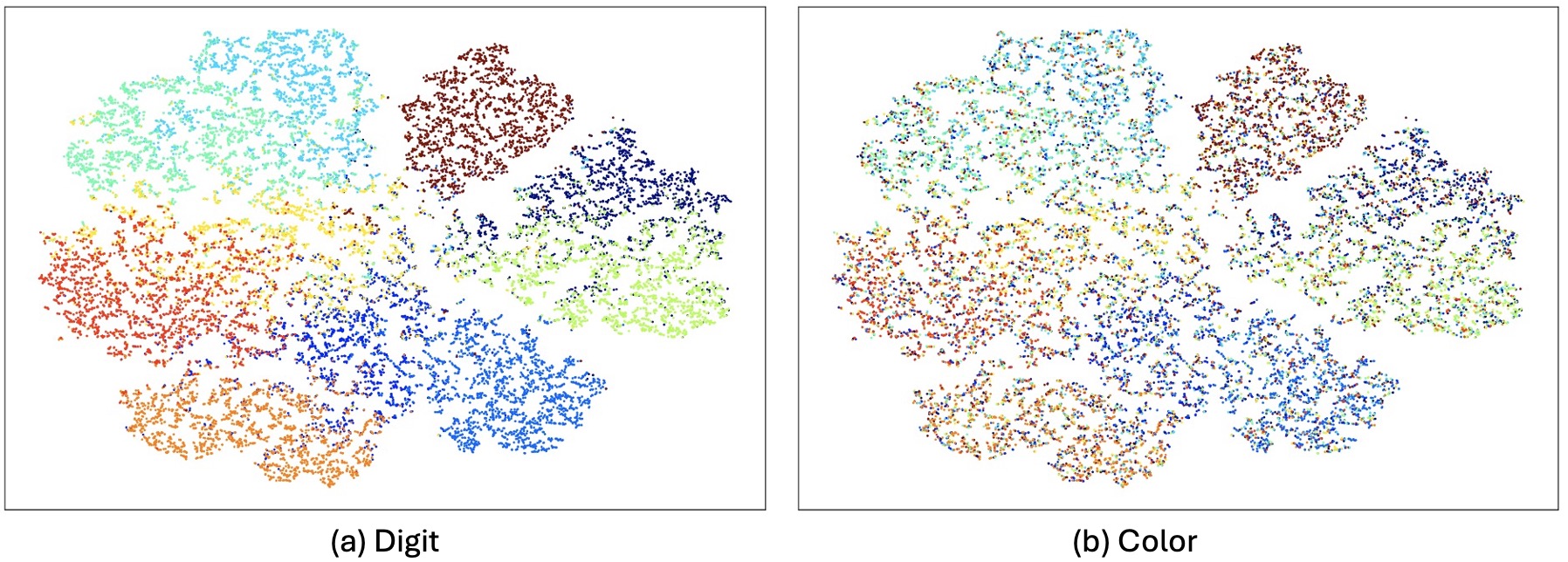}
         \caption{ODVAE~\cite{sarhan2020fairness}}
         \label{fig:sub5}
    \end{subfigure}
    \hfill
    \begin{subfigure}[b]{0.49\linewidth}
         \centering
         \includegraphics[width=\linewidth]{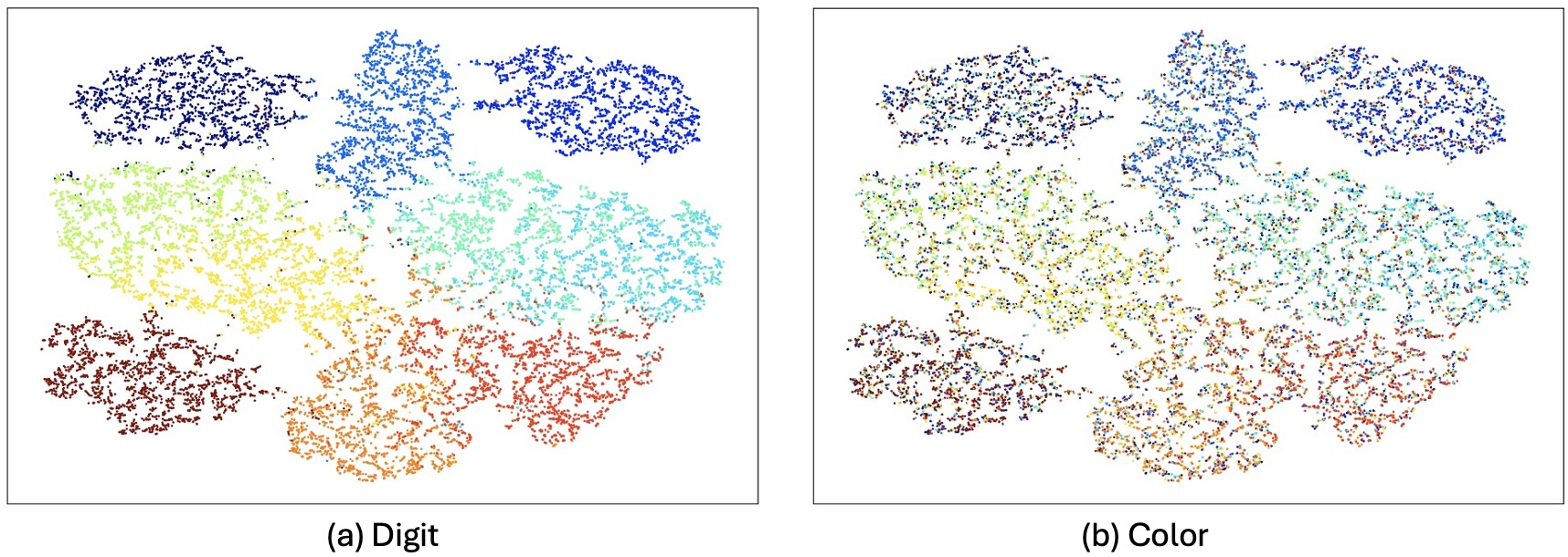}
         \caption{FairDisCo~\cite{liu2022fair}}
         \label{fig:sub6}
    \end{subfigure}
    
    \vspace{1em}

    \begin{subfigure}[b]{0.49\linewidth}
         \centering
         \includegraphics[width=\linewidth]{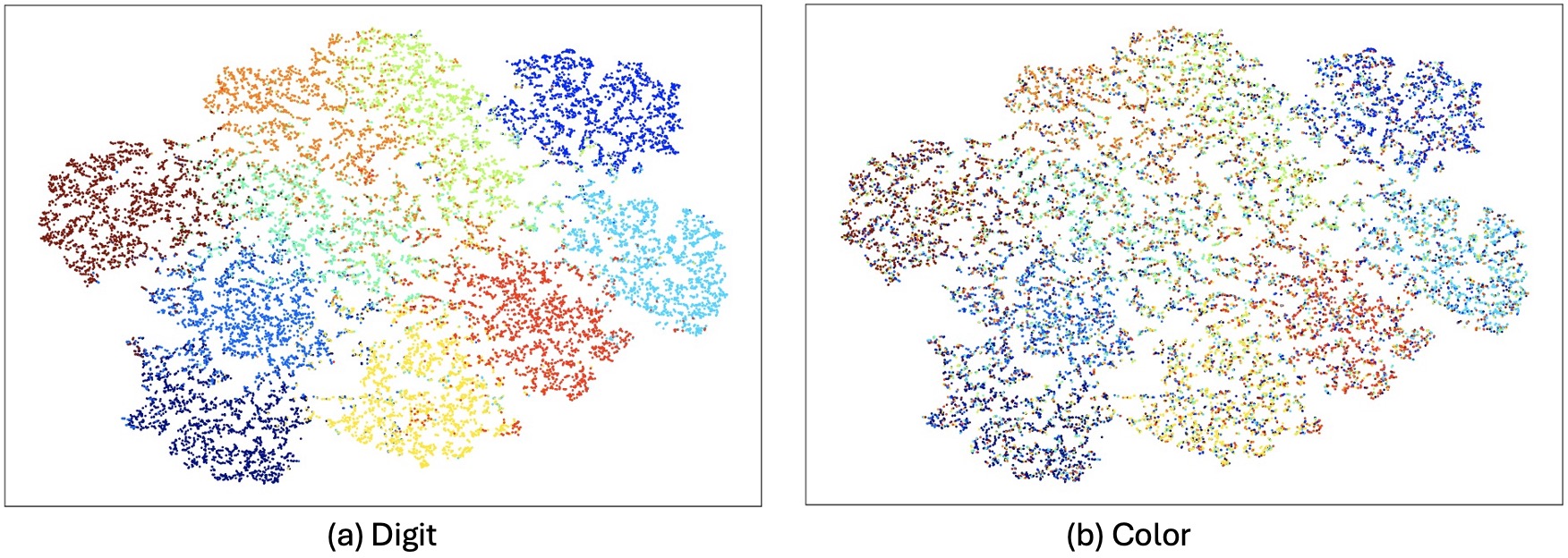}
         \caption{FairFactorVAE~\cite{liu2023fair}}
         \label{fig:sub7}
    \end{subfigure}
    \hfill
    \begin{subfigure}[b]{0.49\linewidth}
         \centering
         \includegraphics[width=\linewidth]{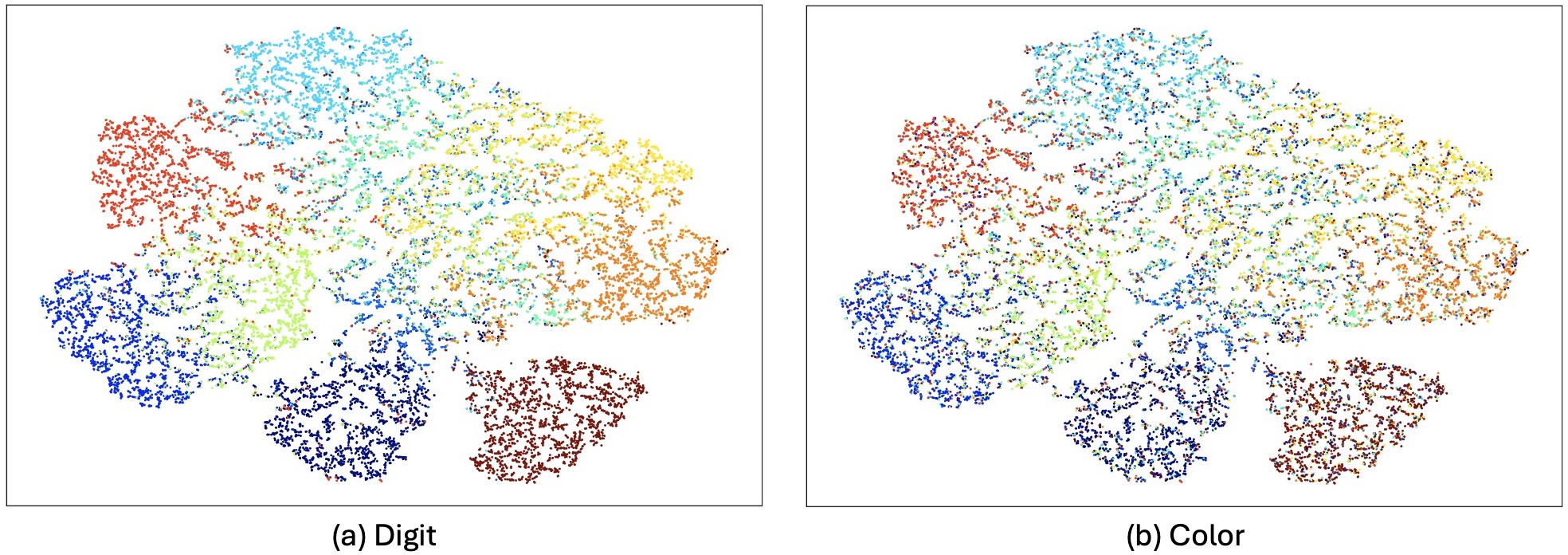}
         \caption{FactorVAE~\cite{kim2018disentangling}}
         \label{fig:sub8}
    \end{subfigure}
    
    \caption{\textbf{t-SNE visualization of the target code from the test set for each method}. Left subfigure is colored by Digit; right subfigure is colored by Color.}

    \label{fig:t-SNE}
\end{figure*}

\subsection{t-SNE Visualization}
To better understand the distribution and disentanglement of the learned representation, we present a t-SNE visualization analysis of the target latent code 
of each method. The visualizations are derived from experiments in Section 4.1 of the main paper where the model is trained on a biased color MNIST dataset and tested on an unbiased color MNIST dataset. Each figure consists of two subfigures: the left subfigure is colored according to the Digit attribute (target attribute), and the right subfigure is colored according to the Color attribute (sensitive attribute). Clear and distinct clustering in the left subfigure indicates that the model has learned a robust and discriminative representation of the target attribute, thereby enhancing its recognizability. Conversely, if the right subfigure exhibits discernible color clusters, it suggests a correlation between the target and sensitive attributes, indicating weaker disentanglement performance. A uniform color distribution in the right subfigure, however, confirms that the sensitive information has been effectively filtered out.

This visualization in Figure~\ref{fig:t-SNE} demonstrates that our proposed method effectively disentangles the learned representation. The target attribute (Digit) exhibits distinct, well-separated clusters with clear classification boundaries and a pure distribution, while the sensitive attribute (Color) is uniformly distributed and unrecognizable. This confirms that our method achieves superior separation of the target attribute without introducing unwanted bias from the sensitive attribute.

\subsection{Text-to-Image Editing}
To further validate the capability and explore the applicability of our method, we integrated it as an adaptor on top of a pre-trained, frozen CLIP image encoder \cite{radford2021learningtransferablevisualmodels} and trained it on datasets including CelebA~\cite{liu2015deep} and Facet~\cite{gustafson2023facetfairnesscomputervision} to enhance fairness in vision-language tasks. Table~\ref{tb:sup_CLIP_celebA} and Table~\ref{tb:sup_CLIP_facet} present the experimental results on CelebA and Facet, respectively. These results demonstrate that our approach significantly improves fairness without compromising performance compared to the linear probing baseline (ERM), underscoring its potential for a range of vision-language tasks such as search and image retrieval with fairness considerations. 

In Table~\ref{tb:sup_CLIP_facet}, we also present an ablation study that examines the contribution of each loss term, as detailed in the last three rows. Each row represents training using only $L_{ELBO}$, without $L_{CMI}$, and without $L_{LRI}$, respectively. 

\begin{table}[!htbp]
\centering
\caption{\textbf{Performance of CLIP(ViTB/32) on CelebA dataset.}}
\label{tb:sup_CLIP_celebA}
\renewcommand{\arraystretch}{0.9} 
\setlength{\tabcolsep}{4pt} 
\begin{adjustbox}{width=0.7\columnwidth,center}
\begin{tabular}{l||ccc} 
\Xhline{2\arrayrulewidth}
\rowcolor{gray!20} \textbf{Method} & Acc $\uparrow$ & EOD $\downarrow$ & DP $\downarrow$ \\ 
\Xhline{1.5\arrayrulewidth}
Zero-shot & 0.857 & 0.834 & 0.715 \\ 
Linear prob & 0.918 & 0.924 & 0.837 \\ 
CAD-VAE & \textbf{0.921} & \textbf{0.037} & \textbf{0.133} \\ 
\Xhline{2\arrayrulewidth}
\end{tabular}
\end{adjustbox}
\vspace{-1.2em}
\end{table}
\begin{table*}[htbp]
\centering
\caption{\textbf{Performance of CLIP(ViTB/32) on Facet dataset.} WG: Worst Group, Gap: Difference between WG and Avg.}
\label{tb:sup_CLIP_facet}
\renewcommand{\arraystretch}{0.9} 
\setlength{\tabcolsep}{3pt} 
\small 
\begin{adjustbox}{width=0.7\textwidth,center}
\begin{tabular}{l||ccc|ccc} 
\Xhline{2\arrayrulewidth}
\rowcolor{gray!20}  
\textbf{Method} & \multicolumn{3}{c|}{\textbf{Top-1 Acc. (\%)}} & \multicolumn{3}{c}{\textbf{Top-3 Acc. (\%)}} \\  
\rowcolor{gray!20}  
& WG $\uparrow$ & Avg $\uparrow$ & Gap $\downarrow$ & WG $\uparrow$ & Avg $\uparrow$ & Gap $\downarrow$ \\ 
\Xhline{1.5\arrayrulewidth} 
Zero-shot   & 2.79 & 53.45 & 50.66 & 15.31 & 76.79 & 61.48 \\ 
Linear prob & 1.17 & 65.46 & 64.29 & 1.79  & 85.34 & 83.55 \\ 
CAD-VAE       & \textbf{69.97} & \textbf{70.54} & \textbf{0.57} & \textbf{85.36} & \textbf{85.95} & \textbf{0.59} \\ 
(Abl.) $\mathcal{L}_{\mathrm{ELBO}}$ & 16.34 & 67.27 & 50.93 & 37.24 & 86.96 & 49.72 \\ 
(Abl.) w/o $\mathcal{L}_{\mathrm{CMI}}$ & 20.43 & 67.67 & 47.24 & 25.41 & 91.12 & 65.71 \\ 
(Abl.) w/o $\mathcal{L}_{\mathrm{LRI}}$ & 24.03 & 64.19 & 40.16 & 29.17 & 87.42 & 58.25 \\ 
\Xhline{2\arrayrulewidth}
\end{tabular}
\end{adjustbox}
\vspace{-1.5em}
\end{table*}

Furthermore, we applied our method in StyleCLIP~\cite{Patashnik_2021_ICCV} as a fair discriminator to address inherent fairness issues, such as career-gender biases, which persist even when an identity preservation loss is employed. As illustrated in Figure~\ref{fig:StyleCLIP}, StyleCLIP~\cite{Patashnik_2021_ICCV} exhibits a bias by correlating the role of ``dancer" with a specific gender. In contrast, our method effectively mitigates this bias while maintaining the efficacy of attribute modification.

\begin{figure}[h]

    \centering
    \begin{subfigure}[b]{0.32\linewidth}
        \centering
        \includegraphics[width=\linewidth]{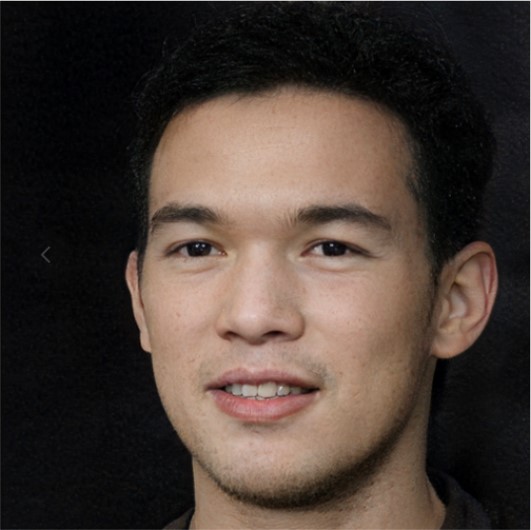}
        \caption{Original}
        \label{fig:sub1}
    \end{subfigure}%
    \hfill
    \begin{subfigure}[b]{0.32\linewidth}
        \centering
        \includegraphics[width=\linewidth]{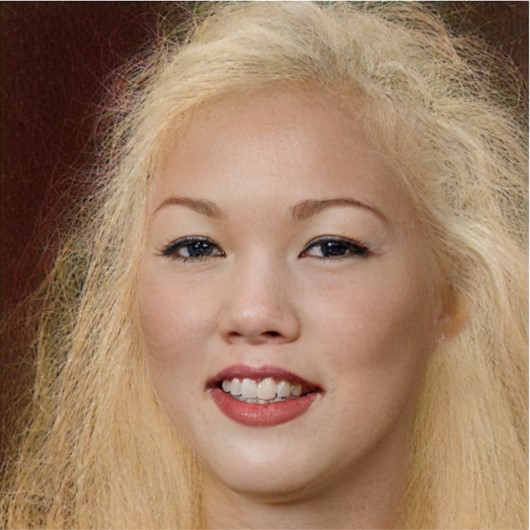}
        \caption{StyleCLIP}
        \label{fig:sub2}
    \end{subfigure}%
    \hfill
    \begin{subfigure}[b]{0.32\linewidth}
        \centering
        \includegraphics[width=\linewidth]{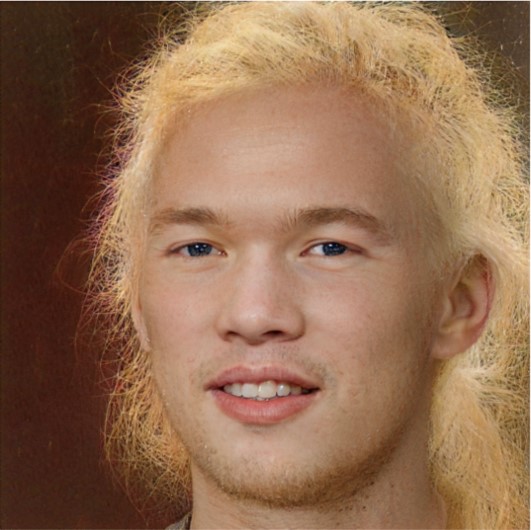}
        \caption{StyleCLIP+}
        \label{fig:sub3}
    \end{subfigure}
    \vspace{-1em}
    \caption{\textbf{Style transfer using StyleCLIP and the CAD-VAE extension.} This example transforms the (a) into ``a dancer with long blonde hair." ``StyleCLIP+" means StyleCLIP + CAD-VAE.}
    \label{fig:StyleCLIP}
    \vspace{-2em}

\end{figure}

\section{Limitation and Future Works}
\label{sec:appendix_section_F}

The limitations of our work can be summarized in two main aspects. First, while our study addresses the inherent trade-off between fairness and performance under certain data biases, fairness violations can arise from a variety of sources. CAD-VAE specifically mitigates the unwanted correlation between sensitive attribute and target attribute, which represents one primary cause of fairness issues. However, in real-world applications, additional factors such as under-representation, intrinsic model bias, and the presence of missing or noisy features often co-occur, exacerbating fairness violations. Future research should aim to extend our theoretical framework to encompass these diverse and often overlapping sources of bias.
Second, our current approach requires the availability of both target and sensitive attribute information during training. In many practical scenarios, acquiring such labeled data can be challenging due to high annotation costs or legal and regulatory restrictions. A promising direction for future work is to develop methods that relax this dependency, potentially through unsupervised or semi-supervised techniques, to learn fair and disentangled representations without explicit reliance on labeled sensitive information.

\clearpage



\end{document}